\documentclass[11pt,a4paper]{article}
\usepackage[T1]{fontenc}
\usepackage{lmodern}
\usepackage[a4paper,margin=1in]{geometry}
\usepackage{amsmath,amssymb}
\usepackage{graphicx}
\usepackage{booktabs}
\usepackage{array}
\usepackage[colorlinks=true,allcolors=blue]{hyperref}
\usepackage{cite}

\newcommand{\tblwidth}{\linewidth}
\title{Matrix-Free Photoacoustic Image Reconstruction via Sensor-Token Self-Attention}
\author{Mary John\textsuperscript{a,*}, Shibili Said\textsuperscript{b}, Imad Barhumi\textsuperscript{b}, Sherzod Turaev\textsuperscript{b}, Mohamed Yahia\textsuperscript{a}\\[6pt]
\small \textsuperscript{a}Abu Dhabi Polytechnic, Abu Dhabi, United Arab Emirates\\
\small \textsuperscript{b}United Arab Emirates University, Al Ain, United Arab Emirates\\[4pt]
\small \textsuperscript{*}Corresponding author: mary.john@actvet.gov.ae}
\date{}

\begin{document}
\maketitle

\begin{abstract}
Photoacoustic tomography (PAT) couples optical absorption contrast with ultrasound resolution, but reconstructing the initial pressure distribution from sparse-view measurements is an ill-posed inverse problem. Iterative compressive-sensing solvers and unrolled deep networks both retain a dependence on the system matrix at inference, leaving real-time clinical use computationally expensive. This paper proposes the Sensor Attention Network (SAN), a Transformer-based architecture that treats each sensor's time series as a token and maps raw measurements directly to the reconstructed image without invoking the system matrix at inference. For training and benchmarking, an analytical k-space H-matrix is constructed and validated against the k-Wave solver, achieving a mean per-sensor Pearson correlation of $\bar{r}=0.919\pm 0.049$. Trained with a vessel-weighted loss on 488 augmented samples and evaluated on 46 held-out samples against ISTA, split-Bregman total variation (SBTV), and learned ISTA (LISTA), SAN attains the highest mean SSIM (0.522) and PSNR (22.09 dB) and the lowest NMSE (0.233). Paired t-tests and Wilcoxon signed-rank tests confirm SAN's superiority over LISTA on PSNR, NMSE, and Pearson correlation at $p<10^{-8}$, and over ISTA and SBTV on all fidelity metrics. By bypassing the H-matrix at inference, SAN reduces reconstruction time by at least an order of magnitude, supporting real-time PAT reconstruction.
\end{abstract}


\vspace{1em}

\section{Introduction}\label{1}
Photoacoustic tomography (PAT) has emerged over the past two decades as one of the most promising hybrid biomedical imaging modalities, combining the rich endogenous optical absorption contrast of haemoglobin, melanin, and lipids with the deep tissue penetration and high spatial resolution of ultrasound \cite{beard2011biomedical, 2}. A nanosecond pulsed laser illumination of biological tissue causes localised heating and thermoelastic expansion, generating broadband acoustic pressure waves that are detected by an array of ultrasound transducers placed around the tissue. The recovered initial pressure distribution provides a quantitative spatial map of optical absorbers and underpins emerging clinical applications including breast cancer screening, vascular imaging, melanoma detection, and functional neuroimaging \cite{3, luke2012biomedical}.

The reconstruction of the initial pressure distribution from the time-resolved sensor signals is governed by the linear acoustic wave equation and, under the assumption of a homogeneous lossless medium, can be expressed compactly as a linear inverse problem $\mathbf{y} = \mathbf{H}\mathbf{x}_{0} + \mathbf{n}$, where $\mathbf{H}$ is the discretised forward (system) operator, $\mathbf{x}_0$ is the vectorized image (gray-scale) pixels and $\mathbf{n}$  is the noise. In practice, $\mathbf{H}$ is severely under-determined or ill-posed because (i) the sensor array geometry is sparse and frequently subtends less than $2 \pi$  around the tissue, (ii) the time-sampling and grid-discretisation introduce numerical artefacts, and (iii) measurement noise compounds at high spatial frequencies \cite{5,6}. Consequently, naive inversion (e.g., back-projection or filtered back-projection) yields reconstructions corrupted by streaking artefacts, smearing, and poor resolution of fine vascular detail \cite{john2022fast}.

To overcome ill-posedness, the PAT community has invested heavily in compressive sensing formulations that exploit the spatial sparsity of vessel-like structures. Iterative algorithms such as Iterative Shrinkage-Thresholding Algorithm (ISTA) and the split-Bregman total-variation (SBTV) method have been shown to outperform classical back-projection in limited-view settings, recovering coarse structure with high spatial correlation \cite{john2022total8,john2023compressive9}. A systematic comparison of these reconstruction algorithms under full-view and limited-view conditions was reported in \cite{john2022fast}, where iterative $l_1$-regularised solvers were shown to be the most robust baselines. Our own comparative analysis \cite{john2022fast} found SBTV to be the most accurate of these classical solvers, but at the cost of a preprocessing step and a per-iteration complexity that scales with the dimensions of $\mathbf{H}$, limiting its use as grid and sensor counts increase. Plug-and-play frameworks that integrate learned denoising priors with classical iterative updates further improved fidelity for limited-sample regimes \cite{john2023plug11}.
More recently, deep learning has been applied to PAT reconstruction in two principal modes \cite{john2023advancing10,shahid2021deep12, antholzer2019deep13,john2025unrolled14}. The first uses convolutional encoder-decoder networks as post-processing stages applied to an initial classical estimate, reducing artefacts but retaining the system-matrix dependence in the front-end. The second is algorithm unrolling, in which a fixed number of iterations of an iterative algorithm is unfolded into a trainable deep network with each layer corresponding to one iteration \cite{li2021deep15}. Unrolled networks preserve the physics of the forward operator while exploiting data-driven priors, and have demonstrated strong reconstruction performance under sparse-view PAT conditions \cite{5,john2023advancing10,john2025unrolled14}. Despite these advances, both paradigms retain repeated $\mathbf{Hx}$ and $\mathbf{H}^{T}$ multiplications at inference time, which for clinically relevant grid resolutions can dominate the per-reconstruction cost \cite{josy2024optimizing16}.
This paper proposes a complementary line of attack: a direct, data-driven mapping from raw sensor measurements to the reconstructed image that completely bypasses the system-matrix at inference. The proposed architecture, named the Sensor Attention Network (SAN), treats each sensor’s full time-series as a token in a Transformer encoder \cite{vaswani2010ls17}. Multi-head self-attention allows every sensor to interact with every other sensor, capturing the non-local acoustic correlations that the analytical H-matrix encodes implicitly. A learned spatial decoder maps the attended sensor representation back to the image domain. In inference, reconstruction reduces to a single forward pass through the network, taking milliseconds on a single GPU. Crucially, the analytical system-matrix is not discarded but repurposed: it encodes the acoustic physics that make the forward problem well-defined, and is used to synthesise physically faithful training data from which the network learns the inverse mapping. The system matrix therefore underpins training, while inference is freed of its per-reconstruction cost.

The principal contributions of this work are summarised as follows:
\begin{enumerate}
\item An analytical k-space forward model is constructed and validated against the k-Wave pseudo-spectral time-domain solver under matched geometry, achieving a mean per-sensor Pearson correlation of $\bar r=0.919\pm0.049$. Two targeted regularisations: k-space apodization and Gaussian temporal damping are introduced and shown by ablation to act synergistically, reducing the energy-normalised mismatch by $49 \%.$
\item The Sensor Attention Network (SAN), a sensor-as-token Transformer encoder followed by a learned spatial decoder, is proposed as a direct (non-iterative, system-matrix-free) reconstruction architecture for PAT.
\item A comprehensive empirical comparison is reported against three representative baselines, ISTA, SBTV, and Learned Iterative Shrinkage Algorithm (LISTA), on the same 46-sample held-out test set, using Structural Similarity Index Measure (SSIM), Peak Signal-to-Noise Ratio (PSNR), Normalized Mean Squared Error (NMSE), and Pearson correlation. The SAN achieves the highest mean SSIM and PSNR and the lowest mean NMSE of the four methods, with paired statistical tests confirming significance at $p<10^{-8}$ for the principal claims.
\end{enumerate} 
The remainder of the paper is organised as follows: Section \ref{2} reviews related work in PAT reconstruction. Section \ref{3} describes the analytical H-matrix construction, its regularisation, and the validation methodology. Section \ref{4} introduces the proposed Sensor Attention Network. Section \ref{5} reports the experimental setup and full numerical comparison. Section \ref{6} discusses interpretation, limitations, and clinical implications. Section \ref{7} concludes the paper and outlines directions for future work. 

\section{Related Works}\label{2}
Reconstruction algorithms for PAT can be broadly grouped into four families: analytical back-projection methods, classical iterative compressive-sensing methods, model-based deep-learning methods, and direct (end-to-end) deep-learning methods. The proposed Sensor Attention Network belongs to the fourth family. We briefly review each below.

\subsection{Analytical and Iterative Compressive-Sensing Reconstruction}\label{2.1}
Analytical reconstruction by universal back-projection \cite{xu2005universal18} provides closed-form image estimates under idealised geometry assumptions, but degrades rapidly under sparse or limited-view sensing. Compressive-sensing-based iterative methods address this limitation by exploiting the sparsity of vessel images in suitable transform domains. ISTA \cite{19} solves an $l_1$-regularised least-squares problem and has been widely applied to PAT \cite{john2022fast}. The SBTV formulation \cite{qin2015alternating20} enforces piecewise-smooth structure on the reconstructed image and has been particularly effective in suppressing high-frequency artefacts arising from sparse sensor arrays. Total-variation-based reconstruction for PAT specifically was studied in \cite{john2022total8}, where the choice of regularisation parameter and stopping criterion were shown to strongly influence reconstruction fidelity. A systematic comparison of full-view and limited-view iterative methods was reported in \cite{john2023compressive9}, demonstrating that no single algorithm dominates across all sensing configurations and that the appropriate method depends on the sparsity geometry. In our earlier comparative work, ISTA-based reconstruction was examined in \cite{john2023advancing10} and split-Bregman total variation in \cite{john2022fast}, establishing the classical iterative baselines used in the present study.

\subsection{Limited-View and Limited-Sample PAT}\label{2.2}
The limited-view regime, in which the sensor array subtends less than the full $2\pi$ around the tissue introduces additional ill-posedness because spatial frequencies aligned with the missing aperture directions cannot be recovered. Compressive sensing techniques tailored to this scenario have been studied in \cite{john2023compressive9}, where iterative $l_1$ solvers were combined with carefully chosen sensor sub-sampling patterns to maintain reconstruction quality at reduced sensor counts. The complementary problem of limited-sample PAT, recovering the initial pressure from a small training corpus, which is the regime relevant to many clinical settings, was addressed in \cite{john2023plug11, venkatakrishnan2013plug21, pp22,awasthi2021dimensionality23}  using a plug-and-play framework that combines a classical iterative solver with a learned denoiser. The handling of missing data in limited-view PAT, where some sensor signals are corrupted or absent, was addressed by a compressive-sensing-driven deep-learning approach in \cite{5}.

\subsection{Model-Based Deep Learning and Algorithm Unrolling}\label{2.3}
Algorithm unrolling \cite{li2021deep15} unfolds a fixed number of iterations of an iterative reconstruction algorithm into a trainable deep network, with each layer corresponding to one iteration. The proximal operator within each layer (typically a soft-threshold for $ L_1$-regularised problems) is replaced by a learnable mapping, while the data-fidelity step retains the forward operator $\mathbf{H}$ and its adjoint $\mathbf{H}^{\mathrm{T}}$ explicitly. The LISTA \cite{19, gregor2010learning24} is the canonical example. Unrolled networks for PAT have been studied in \cite{john2023advancing10}, where the architecture was made intelligible by exposing per-layer thresholds for clinical interpretation, and in \cite{5}, where unrolled networks were applied to breast-cancer-related limited-view PAT data. While these methods offer substantial improvements in reconstruction quality over classical solvers and retain physical interpretability, they share an important inference-time limitation: every layer requires an $\mathbf{H}\mathbf{x}$ and $\mathbf{H}^{\mathrm{T}}$ multiplication (each an $O(MN)$ matrix-vector product), so the per-image computational cost still scales as $K\,O(MN)$, where $K$ is the unrolled depth. For clinically realistic image sizes and sensor counts, this remains a bottleneck for real-time use.

\subsection{Direct (End-to-End) Deep Learning for Image Reconstruction}\label{2.4}
A more recent line of work seeks to avoid the H-matrix at inference entirely, by training a deep network to map sensor measurements directly to the reconstructed image. Convolutional neural networks and U-Net variants have been used in this role \cite{2, john2023advancing10, lan2020net25,hsu2021comparing26, rajendran2022photoacoustic27}, but their treatment of the sensor array as a regular grid does not exploit the nonlocal
acoustic relationships between sensors. 
In a related generative direction, diffusion-based models have recently been
investigated for PAT reconstruction, with comparative studies benchmarking
several such frameworks against classical solvers~\cite{said2026comparative}. More recently, attention-based mechanisms have been explored in PAT reconstruction to improve feature extraction and information fusion from sparse measurements \cite{yin2025optimizationAT1, lai2026attentionAT2}. However, these approaches typically apply attention within learned feature representations rather than explicitly modelling interactions among individual sensors.
The proposed Sensor Attention Network belongs to this family but uses a Transformer-based architecture [28] in which each sensor is represented as a token and global self-attention captures the cross-sensor relationships explicitly. The benefit of this design is two-fold: it preserves the physical role of the sensor array (sensors are unordered entities communicating via the acoustic field) and it enables fully parallel inference without iterative dependence on $\mathbf{H}$.

\subsection{Positioning of the Proposed Work}\label{2.5}
The present work differs from prior compressive-sensing PAT reconstruction in that the H-matrix is not invoked at inference; it differs from prior unrolled-network PAT reconstruction in that no per-layer forward/adjoint operation is required and it differs from prior CNN-based direct reconstruction in that the sensor array is treated as a set of tokens with explicit self-attention rather than as a structured 2-D input. The validated analytical H-matrix from Section \ref{3} is used only at the training-data generation stage to synthesise sensor measurements from phantom images, after which the network learns the inverse mapping without further reference to H.

\section{METHODOLOGY}\label{3} 
\subsection{Problem Formulation}\label{3.1}
In PAT, a short-pulsed laser illuminates a biological tissue sample, inducing a localised temperature rise that generates a broadband acoustic pressure wave through the thermoacoustic effect. The spatial distribution of this initial acoustic pressure, denoted $\mathbf{x}_0 \in \mathbb{R}^{N_x N_y}$ (where $N_x$ and $ N_y$ are the vectorized image grid dimensions in pixels), encodes the optical absorption map of the tissue and constitutes the quantity of interest. A surrounding array of $N_s$ ultrasound transducers records the time-varying pressure field over $N_t$ discrete time steps, producing the measurement vector $\mathbf{y} \in \mathbb{R}^{N_s N_t}$. Under the assumption of a lossless, homogeneous medium with constant speed of sound $c$, the forward problem is linear and may be written compactly as:
\begin{equation}
\mathbf{y} = \mathbf{H}\mathbf{x}_0 + \mathbf{n}
\label{eq:forward_model}
\end{equation}

where $\mathbf{H} \in \mathbb{R}^{N_sN_t \times N_xN_y}$ is the system matrix (hereafter the H-matrix) mapping the vectorised image $\mathbf{x}_0$ to all sensor measurements, and $\mathbf{n}$ represents additive measurement noise. The reconstruction problem estimating $\mathbf{x}_0$ from $\mathbf{y}$ is an ill-posed linear inverse problem, the solution quality of which depends critically on the accuracy of $\mathbf{H}$. The construction and validation of this matrix is the primary concern of this section.

\subsection{Analytical k-Space Forward Model Construction}\label{3.2}
The H-matrix is constructed analytically using a pseudospectral k-space method, which exploits the spatial Fourier decomposition of the initial pressure field. In the k-space domain, the solution to the lossless acoustic wave equation for a two-dimensional homogeneous medium can be expressed at each sensor location $\mathbf{r}_s$ and time $t_n$ as
\begin{equation}
p(\mathbf{r}_s,t_n)
=
\mathbf{W}_{\mathrm{inv}}
\,\mathrm{diag}
\!\left[
\cos\!\left(c|\mathbf{k}|\,t_n\right)
\right]
\mathbf{W}_{\mathrm{fwd}}
\,\mathbf{x}_0
\label{eq:wave_eq}
\end{equation}
where $\mathbf{W}_{\mathrm{fwd}} \in \mathbb{C}^{(N_{\mathrm{out}}^{2}\times N_xN_y)}$ is the two-dimensional forward discrete Fourier transform (DFT) matrix that projects the image onto the $k$-space grid of size $N_{\mathrm{out}}\times N_{\mathrm{out}}$, with $m$ and $l$ indexing the $k$-space; $\mathbf{W}_{\mathrm{inv}} \in \mathbb{C}^{(N_s\times N_{\mathrm{out}}^{2})}$ is the inverse DFT matrix evaluated at the sensor coordinates $\{\mathbf{r}_s\}$; and $|\mathbf{k}|$ denotes the magnitude of the two-dimensional spatial frequency vector 
\begin{align*}
|\mathbf{k}|=\sqrt{k_x^{2}+k_y^{2}} \quad;\quad
k_x=\frac{2\pi m}{N_{\mathrm{out}}d_x} \quad;\quad
k_y=\frac{2\pi l}{N_{\mathrm{out}}d_y}
\end{align*}
where $m \in \left[ \frac{-N_x}{2}, \frac{N_x}{2}-1 \right]$ and
$l \in \left[ \frac{-N_y}{2}, \frac{N_y}{2}-1 \right]$. The cosine term encodes the time evolution of each spectral component under free-space wave propagation. Stacking this expression across all $N_s$ sensors and $N_t$ time steps, the H-matrix takes the factored form

\begin{equation}
\mathbf{H}
=
\mathbf{K}\mathbf{W}_{\mathrm{fwd}}
\label{eq:H_factorisation}
\end{equation}

where $\mathbf{K} \in \mathbb{R}^{N_sN_t \times N_{\mathrm{out}}^{2}}$ is the time-domain propagation kernel, whose row corresponding to sensor $s$ at time $t_n$ is the $s$-th row of $\mathbf{W}_{\mathrm{inv}}$ scaled by the cosine propagation factor $\cos(c|\mathbf{k}|t_n)$. This formulation yields the full system matrix in a single construction pass without iterative simulation, making it computationally efficient for repeated use in the inverse problem.

\subsection{Simulation Grid, Time Array, and Sensor Configuration}\label{3.3}
The imaging domain is discretised on a uniform Cartesian grid of $N_x \times N_y$ pixels with isotropic grid spacing $d_x = d_y$. The choice of $d_x$ governs both the spatial resolution of the reconstructed image and the temporal sampling requirements imposed by the Courant-Friedrichs-Lewy (CFL) stability criterion. For a two-dimensional acoustic problem, the maximum stable time step is

\begin{equation}
d_{t,\max}
=
\frac{d_x}{c\sqrt{2}}
\label{eq:cfl}
\end{equation}

\begin{figure}
  \centering
   \includegraphics[width=70mm]{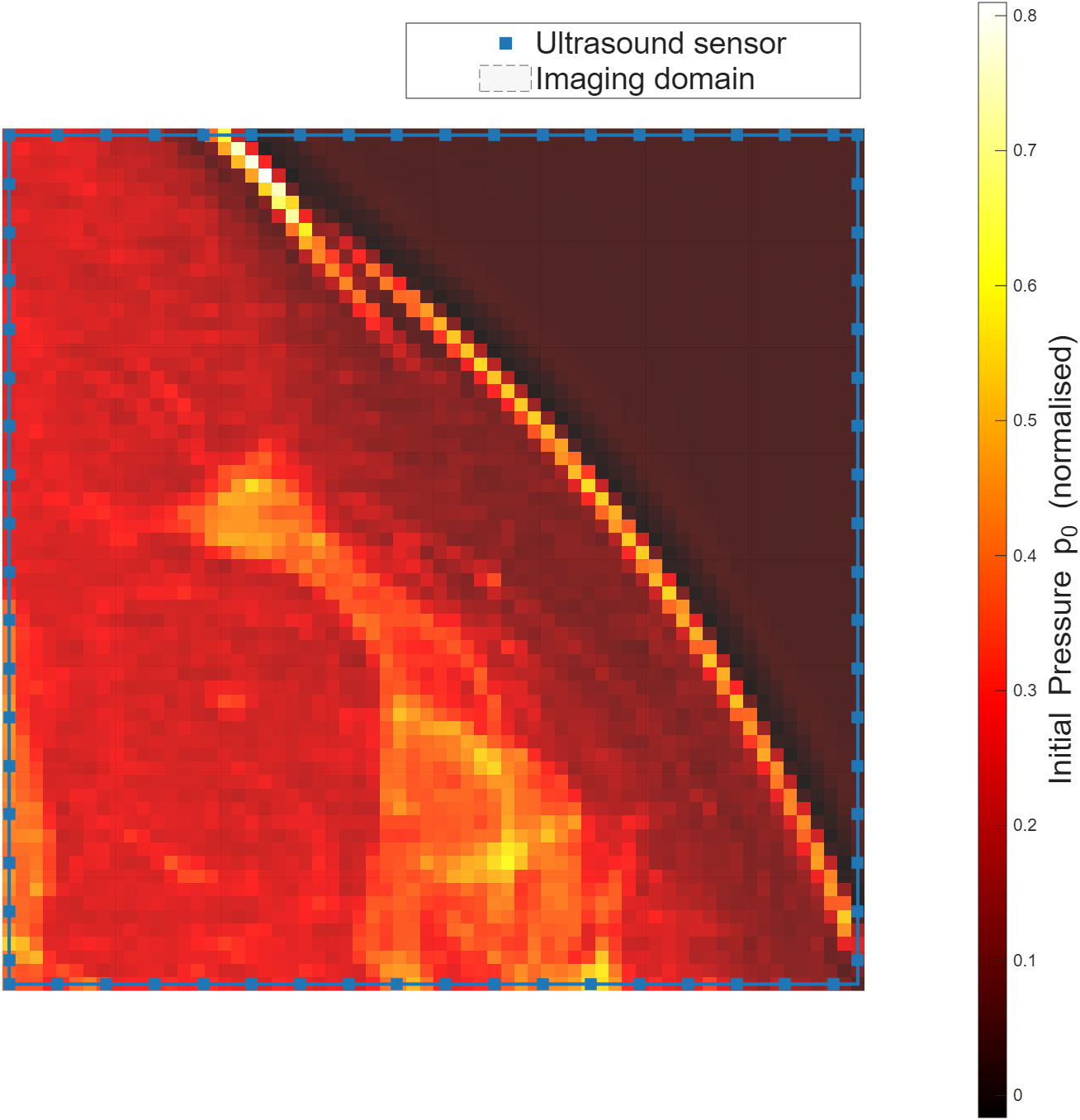}
    \caption{PAT System Geometry: imaging grid (64×64) and square sensor array $(N_{s} = 71)$. $d_{x} = 0.100 mm$.}\label{fig1}
\end{figure}

To ensure that all wavefronts originating within the imaging domain reach the sensor boundary before the simulation terminates, the total measurement duration $T$ and the number of time samples $N_t$ are determined jointly as
\begin{equation}
N_t
=
\frac{T\,\xi}{d_{t,\max}},
\label{eq:Nt}
\end{equation}
where $\xi=1.25$ is a safety factor that provides a temporal margin to prevent signal truncation artefacts. The sensor array consists of $N_s=71$ transducers distributed uniformly along the four edges of a 70$\times$70 boundary grid that encloses the 64×64 imaging domain, with sensors placed at every fourth grid point; the imaging domain, the sensor boundary, and the sensors are all centred within the larger 140$\times$140 k-space simulation grid ($N_{out}$). Because sensors are placed at grid indices 1, 5, …, 69 along each 70-point edge, the index 70 is never sampled; consequently only the shared corner (1,1) is common to two edges, while the remaining three corners fall outside the sampling pattern.
This yields $4 \times 18 - 1 = 71$ unique transducers. 
Sensor Cartesian coordinates were extracted using the k-Wave \texttt{grid2cart()} function and used consistently in both the analytical model and the numerical reference simulation. The complete set of simulation parameters is listed in Table \ref{table1}. The system geometry, including the imaging grid and sensor array configuration, is illustrated in Figure \ref{fig1} and Table \ref{table1}.

\begin{table}[t]
\caption{Simulation Parameters for the Analytical Forward Model ($d_x = 0.100~\mathrm{mm}$)}
\label{table1}
\begin{tabular*}{\tblwidth}{@{\extracolsep{\fill}}llp{4cm}@{}}
\toprule
\textbf{Parameter} & \textbf{Symbol} & \textbf{Value} \\
\midrule
Image grid size  & $N_x \times N_y$   & $64 \times 64$ pixels \\
Grid spacing (selected) & $d_x=d_y$   & $0.100~\mathrm{mm}$ \\
K-space grid size   & $N_{\mathrm{out}}$  & $140 \times 140$ \\
Speed of sound    & $c$    & $1500~\mathrm{m/s}$ \\
Max. stable time step & $d_{t,\max}$  & $4.71\times10^{-8}\mathrm{s}$ \\
Measurement duration  & $T$   & $6~\mu\mathrm{s}$ \\
Time samples (CFL, $\xi=1.5$)    & $N_t$  & $160 $ \\
Number of sensors    & $N_s$  & $71$ \\
Sensor grid boundary & ---  & $70\times 0$, skip = 4 \\
H-matrix dimensions  & $N_sN_t \times N_xN_y$ & $11360\times4096$ \\
PML thickness / coefficient      & ---   & 40 layers, $\alpha = 2$ \\
\bottomrule
\end{tabular*}
\end{table}

\subsection{Regularisation of the Analytical Model}\label{3.4}
In its raw form, the analytical model exhibits two systematic artefacts. First, the truncation of the DFT matrices at the boundaries of the k-space grid introduces Gibbs-like aliasing, manifesting as large-amplitude spikes in the computed sensor signals. Second, without any temporal envelope, the model produces non-physical oscillations at late time samples that have no counterpart in a physically bounded measurement. Two targeted regularisations are applied to suppress these artefacts as stated below:
\begin{enumerate}
\item K-Space Apodization (Windowing): A two-dimensional window function $w(\mathbf{k})$ is applied element-wise to both the forward and inverse DFT matrices prior to construction of $\mathbf{H}$:
\begin{subequations}
\begin{align}
\mathbf{W}_{\mathrm{fwd}}
&\leftarrow
\mathbf{W}_{\mathrm{fwd}}
\odot
w(\mathbf{k}),
\\
\mathbf{W}_{\mathrm{inv}}
&\leftarrow
\mathbf{W}_{\mathrm{inv}}
\odot
w(\mathbf{k})^{T},
\end{align}
\end{subequations}
where $\odot$ denotes the Hadamard (element-wise) product. The window tapers the DFT coefficients smoothly to zero at the k-space boundary, suppressing Gibbs-like aliasing without introducing significant spectral leakage. Three standard apodization functions are candidates for this role: the Hann window, which provides a smooth cosine-squared taper and is the conventional choice for k-space apodization in spectral methods; the Hamming window, which raises the minimum sidelobe level relative to Hann by adding a constant offset; and the Blackman window, which achieves lower sidelobe levels at the cost of a wider main lobe. 
\item Gaussian Temporal Damping:  A Gaussian envelope g(t) is applied to the cosine propagation kernel to enforce realistic signal decay within the measurement window:
\begin{equation}
g(t)
=
\exp\!\left(
-\left(\frac{t}{\tau}\right)^2
\right),
\quad 
\tau = 0.9T 
\label{eq:damping}
\end{equation}

where $\tau$ is the damping time constant (measured in seconds), set to 90\% of the total measurement duration $T$. This ensures the analytical signal decays smoothly before the end of the measurement window, preventing the non-physical late-time oscillations that arise in the undamped model.

\end{enumerate}

\subsection{Numerical Validation via k-Wave}\label{3.5}
The analytical H-matrix is validated by comparing the predicted sensor signals $\mathbf{H}\mathbf{x_{0}}$ against those produced by the k-Wave pseudo-spectral time-domain solver \cite{treeby2010k29}, which numerically integrates the full linearised acoustic wave equation. k-Wave simulations were executed with a perfectly matched layer (PML) of 40 grid points on each side of the domain and an absorption coefficient $\alpha=2$, using an identical spatial grid, time array, sensor configuration, and input pressure image $\mathbf{x}_{0}$ as the analytical model. The degree of agreement between the two models for sensors is quantified by the Pearson correlation coefficient:
\begin{equation}
r_s
=
\frac{\mathrm{Cov}\!\left({\mathbf{y}}_s^{{an}},{\mathbf{y}}_s^{{kw}}\right)}
{\sigma\!\left({\mathbf{y}}_s^{{an}}\right)\cdot
\sigma\!\left({\mathbf{y}}_s^{{kw}}\right)}
\label{eq:8}
\end{equation}

where ${\mathbf{y}}_s^{{an}}$ and ${\mathbf{y}}_s^{{kw}}$ denote the analytical and k-Wave time-series at sensor $s$, respectively. A value of $r_s$ close to unity indicates that the analytical model faithfully reproduces the temporal waveform structure predicted by the numerical simulation. The mean correlation across all $N_s$ sensors, $\bar{r} = \frac{1}{N_s}\sum r_s$, serves as the primary figure of merit for model accuracy.

\subsection{Grid Spacing Study}\label{3.6}

The imaging grid spacing $d_x$ is a critical design parameter because it sets
a direct trade-off between spatial resolution and forward-model fidelity. A
finer grid resolves finer image detail but, through the CFL condition of
\eqref{eq:cfl}, requires a proportionally larger number of time samples $N_t$,
which extends the temporal window over which the analytical and numerical
models must remain in agreement. Since the analytical model assumes free-space
propagation whereas k-Wave enforces absorbing boundary conditions through the
PML, the mismatch between the two accumulates in the late-time signal tail, and
the extent of that tail grows with $N_t$. A larger $d_x$ therefore reduces
model mismatch at the cost of resolution, while a smaller $d_x$ improves
resolution but degrades agreement with the reference.

To select an appropriate value, three candidate grid spacings,
$d_x = 0.075$, $0.100$, and $0.125$~mm, were evaluated against the k-Wave
reference under otherwise identical conditions. For each spacing the full
$H$-matrix and the corresponding k-Wave signals were generated, and the
per-sensor Pearson correlation was computed as the figure of merit, summarised
by its mean, its minimum across the sensor array, and the fraction of sensors
exceeding the acceptance threshold of $r = 0.80$. The spacing yielding the best
agreement on these measures was carried forward to all subsequent experiments.


\section{Proposed Method}\label{4}
The classical iterative and learning-based reconstruction approaches retain a dependence on the H-matrix at inference time, either through repeated matrix-vector products or embedded forward operators. A complementary data-driven approach is proposed in this paper, in which a neural network learns the mapping from sensor measurements $\mathbf{y}$ directly to the reconstructed image $\mathbf{x}^*$. The proposed architecture, referred to as the Sensor Attention Network (SAN), treats each sensor’s time series as a token in a Transformer encoder, allowing every sensor to attend to all others and capture the non-local acoustic relationships encoded in the PAT forward model.
The steps that are adopted for the mapping of sensor data $\mathbf{y}$, to the reconstructed image $\mathbf{x}^*$ are as listed below:

\subsection{Input Projection}\label{4.1}
The sensor measurement $\mathbf{y}$ is reshaped into a matrix $\mathbf{Y}\in\mathbb{R}^{N_t\times N_s}$, so that each column $\mathbf{y}_s\in\mathbb{R}^{N_t}$ is the time series recorded by sensor $s$, for $s = 1,\ldots,N_s$. Each time series is mapped to a $d_{\mathrm{model}}$-dimensional token embedding by a shared linear projection,

\begin{equation}
\mathbf{z}_s
=
\mathbf{W}_{\mathrm{proj}}\mathbf{y}_s
+
\mathbf{b}_{\mathrm{proj}}
\label{eq:token_projection}
\end{equation}
where $\mathbf{W}_{\mathrm{proj}}$ is a learnable weight matrix of size $d_{\mathrm{model}}\times N_t$ and $\mathbf{b}_{\mathrm{proj}}$ is a learnable bias vector of size $d_{\mathrm{model}}$. $\mathbf{z}_s$ is the resulting token embedding for sensor $s$. All token embeddings are assembled column-wise into the matrix $\mathbf{Z}=[\mathbf{z}_1,\ldots,\mathbf{z}_{N_s}]$.

\subsection{Positional Encoding}\label{4.2}
Because the Transformer encoder treats its input as an unordered set, sinusoidal positional encodings \cite{vaswani2017attention28} are added to each token to convey sensor ordering. For sensor index $s$ and embedding dimension index $k$ ($k = 0,1,2,\ldots,d_{\mathrm{model}}-1$), the encoding is defined as

\begin{subequations}
\begin{align}
\mathrm{PE}(s,k)
&=
\sin\!\left(
\frac{s}{10000^{k/d_{\mathrm{model}}}}
\right),
\quad \text{when $k$ is even}
\label{eq:10a}
\\
\mathrm{PE}(s,k)
&=
\cos\!\left(
\frac{s}{10000^{(k-1)/d_{\mathrm{model}}}}
\right),
\quad \text{when $k$ is odd}
\label{eq:10b}
\end{align}
\end{subequations}

The denominator $10000^{k/d_{\mathrm{model}}}$ grows from 1 (at $k = 0$) to 10000 (at $k = d_{\mathrm{model}}$), so that different dimensions oscillate at different frequencies: low-index dimensions change rapidly with $s$ and encode fine position differences, while high-index dimensions change slowly and encode coarse spatial layout. Each token is updated as $\mathbf{z}_s\leftarrow\mathbf{z}_s+\mathrm{PE}(s,:)$, where $\mathrm{PE}(s,:)$ denotes the full $d_{\mathrm{model}}$-dimensional encoding vector for sensor $s$.

\subsection{Multi-Head Self-Attention Encoder}\label{4.3}
The position-encoded token matrix $\mathbf{Z}$ is then processed by $L$ stacked Transformer encoder blocks. Each block applies a multi-head self-attention (MHSA) sub-layer followed by a position-wise feed-forward sub-layer, with a residual connection and layer normalisation after each sub-layer.

In the MHSA sub-layer, $A$ attention heads are computed in parallel. For each head $h\in\{1,\ldots,A\}$, the token matrix $\mathbf{Z}$ is projected into a query matrix $\mathbf{Q}_h$, a key matrix $\mathbf{K}_h$, and a value matrix $\mathbf{V}_h$, each of per-head dimension $d_{\mathrm{head}} = d_{\mathrm{model}}/A$, via

\begin{equation}
\mathbf{Q}_h = \mathbf{W}^{(Q_h)}\mathbf{Z},
\quad
\mathbf{K}_h = \mathbf{W}^{(K_h)}\mathbf{Z},
\quad
\mathbf{V}_h = \mathbf{W}^{(V_h)}\mathbf{Z},
\label{eq:11}
\end{equation}
where $\mathbf{W}^{(Q_h)}$, $\mathbf{W}^{(K_h)}$, $\mathbf{W}^{(V_h)} \in \mathbb{R}^{d_{\mathrm{head}}\times d_{\mathrm{model}}}$ are learnable projection matrices. The query $\mathbf{Q}_h$ encodes what each sensor is searching for, the key $\mathbf{K}_h$ encodes what each sensor offers, and the value $\mathbf{V}_h$ holds the content to be aggregated. The attention output for head $h$ is

\begin{equation}
\mathrm{head}_h = \mathrm{softmax}
\!\left(
\frac{\mathbf{Q}_h\mathbf{K}_h^{T}}
{\sqrt{d_{\mathrm{head}}}}
\right)
\mathbf{V}_h
\label{eq:12}
\end{equation}
where the product $\mathbf{Q}_h\mathbf{K}_h^{T}$ is an $N_s\times N_s$ matrix of raw attention scores, with entry $(i,j)$ measuring the relevance of sensor $j$ to sensor $i$. Division by $\sqrt{d_{\mathrm{head}}}$ prevents scores from becoming excessively large before the softmax operation \cite{19}, which is applied row-wise so that the attention weights for each sensor sum to one. The outputs of all $A$ heads are concatenated and projected back to $d_{\mathrm{model}}$ dimensions via a learnable output matrix $\mathbf{W}_0$ (of size $d_{\mathrm{model}}\times d_{\mathrm{model}}$). A residual connection is added and the result is normalised,

\begin{equation}
\mathbf{Z}
\leftarrow
\mathrm{LayerNorm}
\!\left(
\mathbf{Z}
+
\mathbf{W}_0
\left[
\mathrm{head}_1;
\mathrm{head}_2;
\ldots;
\mathrm{head}_A
\right]
\right)
\label{eq:13}
\end{equation}
where $\mathrm{LayerNorm}$ standardises each token vector to zero mean and unit variance across its $d_{\mathrm{model}}$ dimensions, followed by a learnable scale and shift. The semicolon denotes row-wise concatenation. The residual addition ensures that attention learns only the correction to the existing token representation, preserving information and stabilising gradient flow.

\subsection{Position-Wise Feed-Forward Sub-layer}\label{4.4}

Following the MHSA sub-layer, each token is passed independently through a two-layer feed-forward network (FFN),

\begin{equation}
\mathrm{FFN}(\mathbf{z}_s) = \mathbf{W}_2
\cdot
\mathrm{ReLU}
\!\left(
\mathbf{W}_1\mathbf{z}_s+\mathbf{b}_1
\right)
+
\mathbf{b}_2
\label{eq:14}
\end{equation}
where $\mathbf{W}_1$ (of size $d_{\mathrm{ff}}\times d_{\mathrm{model}}$) expands each token from $d_{\mathrm{model}}$ to a hidden dimension $d_{\mathrm{ff}}$, $\mathrm{ReLU}(\alpha)=\max(0,\alpha)$ is a non-linear activation applied element-wise, and $\mathbf{W}_2$ (of size $d_{\mathrm{model}}\times d_{\mathrm{ff}}$) projects back to $d_{\mathrm{model}}$ dimensions. The same residual-plus-LayerNorm operation as in (\ref{eq:13}) follows. Equations (\ref{eq:11}) to (\ref{eq:14}) together define one complete encoder block; this block is applied $L$ times in sequence.

\subsection{Spatial Decoder}\label{4.5}
After $L$ encoder blocks, the token matrix $\mathbf{Z}$ (of size $d_{\mathrm{model}}\times N_s$) is flattened column-wise into a single vector $\mathbf{z}_{\mathrm{flat}}$ of length $d_{\mathrm{model}}N_s$, which is decoded into the reconstructed image through three fully connected layers,

\begin{subequations}\label{eq:15}
\begin{align}
\mathbf{h}_1
&=
\mathrm{ReLU}
\!\left(
\mathbf{W}_{d_1}\mathbf{z}_{\mathrm{flat}}
+
\mathbf{b}_{d_1}
\right),
\label{eq:15a}
\\
\mathbf{h}_2
&=
\mathrm{ReLU}
\!\left(
\mathbf{W}_{d_2}\mathbf{h}_1
+
\mathbf{b}_{d_2}
\right),
\label{eq:15b}
\\
\mathbf{x}^{*}
&=
\mathrm{sigmoid}
\!\left(
\mathbf{W}_{d_3}\mathbf{h}_2
+
\mathbf{b}_{d_3}
\right),
\label{eq:15c}
\end{align}
\end{subequations}

where $\mathbf{W}_{d_1}$, $\mathbf{W}_{d_2}$, and $\mathbf{W}_{d_3}$ are learnable weight matrices; $\mathbf{h}_1$ and $\mathbf{h}_2$ are intermediate hidden vectors and $\mathrm{sigmoid}(\alpha)=\frac{1}{1+e^{-\alpha}}$ maps each output to $[0,1]$ to match the normalised image range. The final output $\mathbf{x}^{*}$ is reshaped to the desired image size.

\subsection{Training Objective}\label{4.6}
\begin{figure}[h]
\centering
\includegraphics[width=\textwidth]{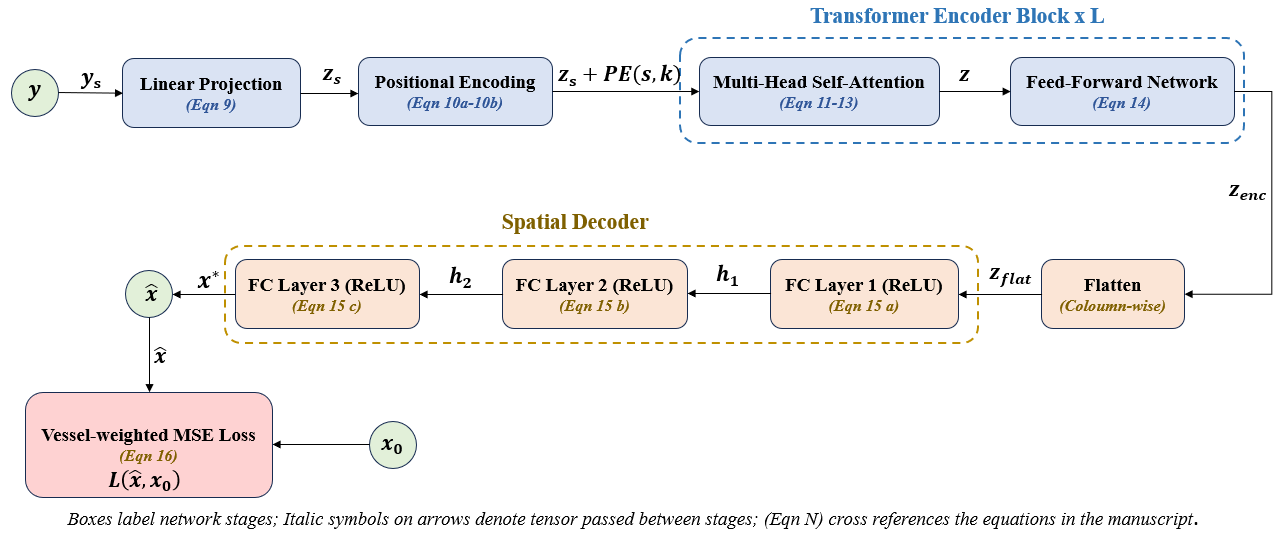}
\caption{SAN Pipeline}
\label{fig2}
\end{figure}

All learnable parameters are jointly optimised by minimising a vessel-weighted mean squared error loss. Since vessel pixels occupy a small fraction of the image area relative to the background, a higher penalty weight is assigned to them. Denoting the ground truth phantom as $\mathbf{x}_0$ and the network prediction as $\mathbf{x}^{*}$, the loss is defined as

\begin{equation}
L
=
\sum_{i}
w_i
\left(
\mathbf{x}_i^{*}
-
\mathbf{x}_{0,i}
\right)^2,
\label{eq:16}
\end{equation}

where $w_i=8$ for vessel pixels (defined as those with $x_{0,i}>0.1$) and $w_i=1$ for background pixels (vessel mask is computed per sample from the ground-truth phantom). The process included in the proposed method is summarized in Figure \ref{fig2}.

\subsection{Dataset Preparation}\label{4.7}
A total of 191 images were downloaded from Radiopaedia.org \cite{radiopaedia}. To mitigate overfitting given the limited sample count, data augmentation was applied by translating each phantom by small integer offsets of 2 to 3 pixels in the horizontal, vertical, and diagonal directions, and recomputing the sensor signal for each translated phantom. Of the 191 source phantoms, 44 were held out for validation and 46 for testing (neither augmented); the remaining training phantoms were augmented by translation to 488 samples, giving 578 images in total.

\section{Results}\label{5}
The experiments reported in this section are organised into five parts. Section \ref{5.1} validates the analytical H-matrix by comparing its predictions against the k-Wave numerical reference. Section \ref{5.2} presents the empirical selection of the k-space apodization window, the Gaussian damping time constant, and the imaging grid spacing. Section \ref{5.3} establishes baseline reconstruction performance using the classical iterative methods ISTA and split-Bregman total variation (SBTV). Section \ref{5.4} reports the training behaviour and validation performance of the proposed Sensor Attention Network. Section \ref{5.5} provides a direct quantitative and qualitative comparison of all four approaches. Throughout, reconstruction quality is evaluated using the structural similarity index measure (SSIM), PSNR, NMSE and Pearson correlation $\bar{r}_{img}$, computed on normalised images in the range [0, 1].

\subsection{Forward Model Validation}\label{5.1}
\begin{figure}
\centering
\includegraphics[width=1\textwidth]{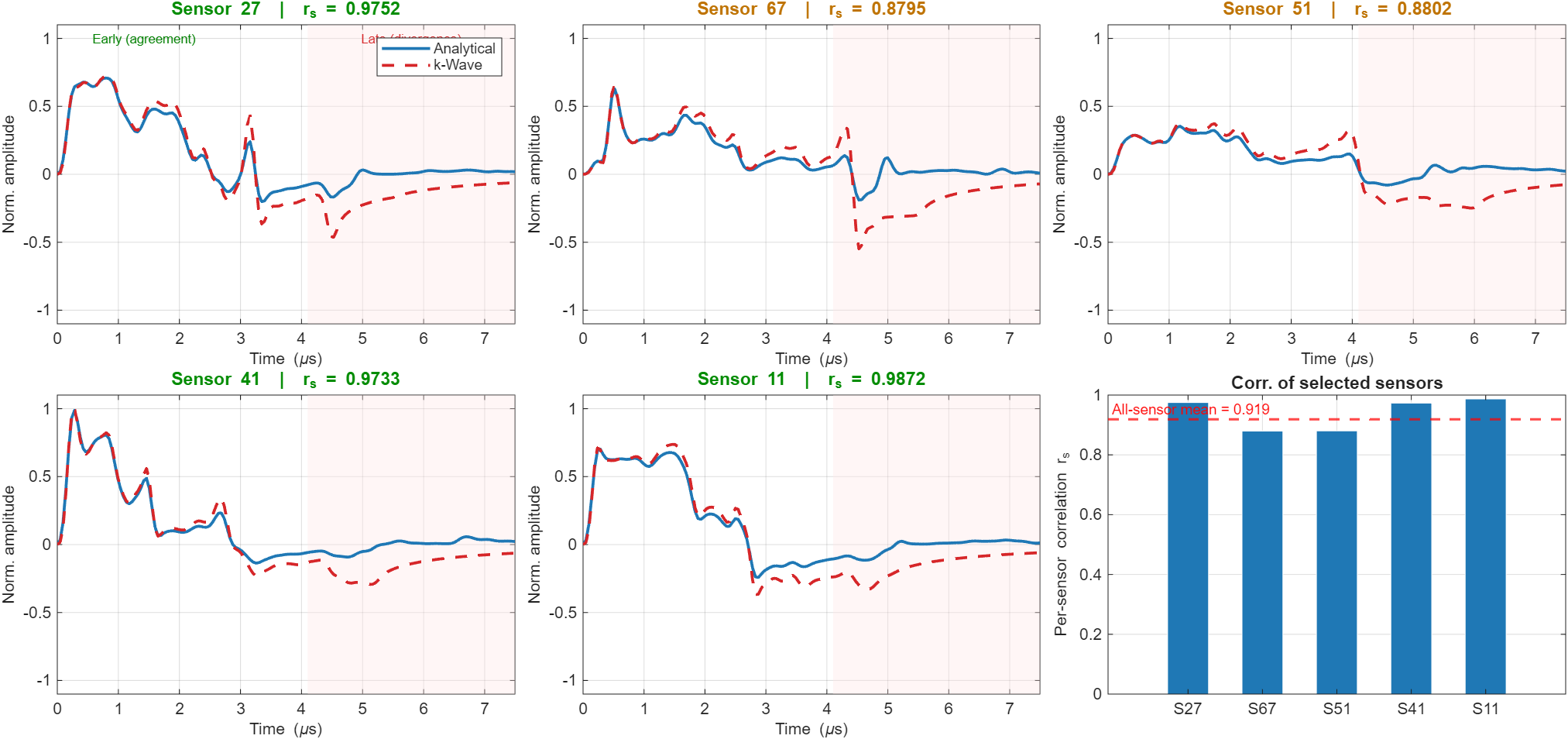}
\caption{Waveform overlays for five randomly selected sensors and their corresponding correlation coefficients.}
\label{fig3}
\end{figure}

The analytical H-matrix was validated by comparing the sensor signals it predicts against those produced by the k-Wave pseudospectral solver under identical imaging conditions. 
For each sensor $s = 1, \ldots$, $N_s$, the Pearson correlation coefficient
$r_s$ between the normalised analytical and k-Wave time series was computed
as defined in (\ref{eq:8}). Figure \ref{fig3} shows representative waveform overlays for five randomly selected sensors, illustrating the close agreement between the two models across the full measurement window of $N_t=160$ time samples.

\begin{figure}
\centering
\includegraphics[width=0.9\textwidth]{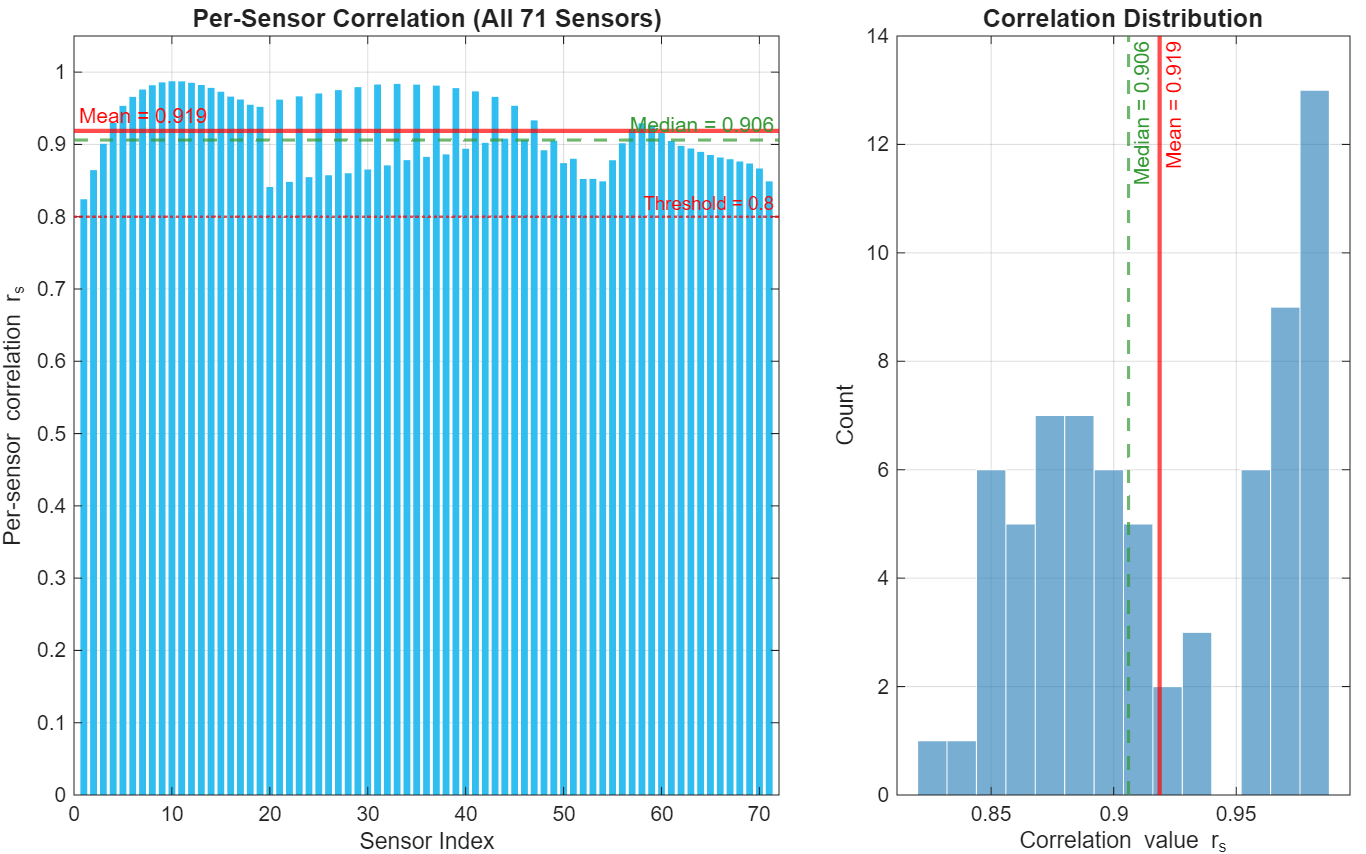}
\caption{Correlation histogram and distribution.}
\label{fig4}
\end{figure}

The mean correlation across all sensors was $\bar{r}=0.919\pm0.049$, with all 71 sensors exceeding $r_s=0.80$. The full per-sensor correlation profile is shown in Figure \ref{fig4}, together with the correlation histogram and summary statistics. The consistently high correlation across the entire sensor array confirms that the analytical model faithfully reproduces the temporal waveform structure of the k-Wave reference, and that of the H-matrix is a reliable and computationally efficient substitute for repeated numerical simulation in the training data generation stage. 
The small residual discrepancy $1-\bar{r} \approx 0.081$ is attributable to the boundary mismatch between the two models: the analytical formulation assumes free-space propagation, whereas k-Wave enforces absorbing boundary conditions via the perfectly matched layer. This discrepancy accumulates preferentially in the late-time signal tail and is effectively suppressed by the Gaussian temporal damping applied during H-matrix construction, as discussed in Section 5.2.

\subsection{Regularisation and Grid Spacing Selection}\label{5.2}
This section reports the empirical selection of the three key design parameters of the analytical forward model: the k-space apodization window, the Gaussian temporal damping constant, and the imaging grid spacing. Each parameter was evaluated independently against the k-Wave reference using Pearson correlation and NMSE as figures of merit, and the optimal configuration was carried forward to all subsequent experiments.

\subsubsection{K-Space Apodization Window}\label{5.2.1}

\begin{figure}
\centering
\includegraphics[width=0.9\textwidth]{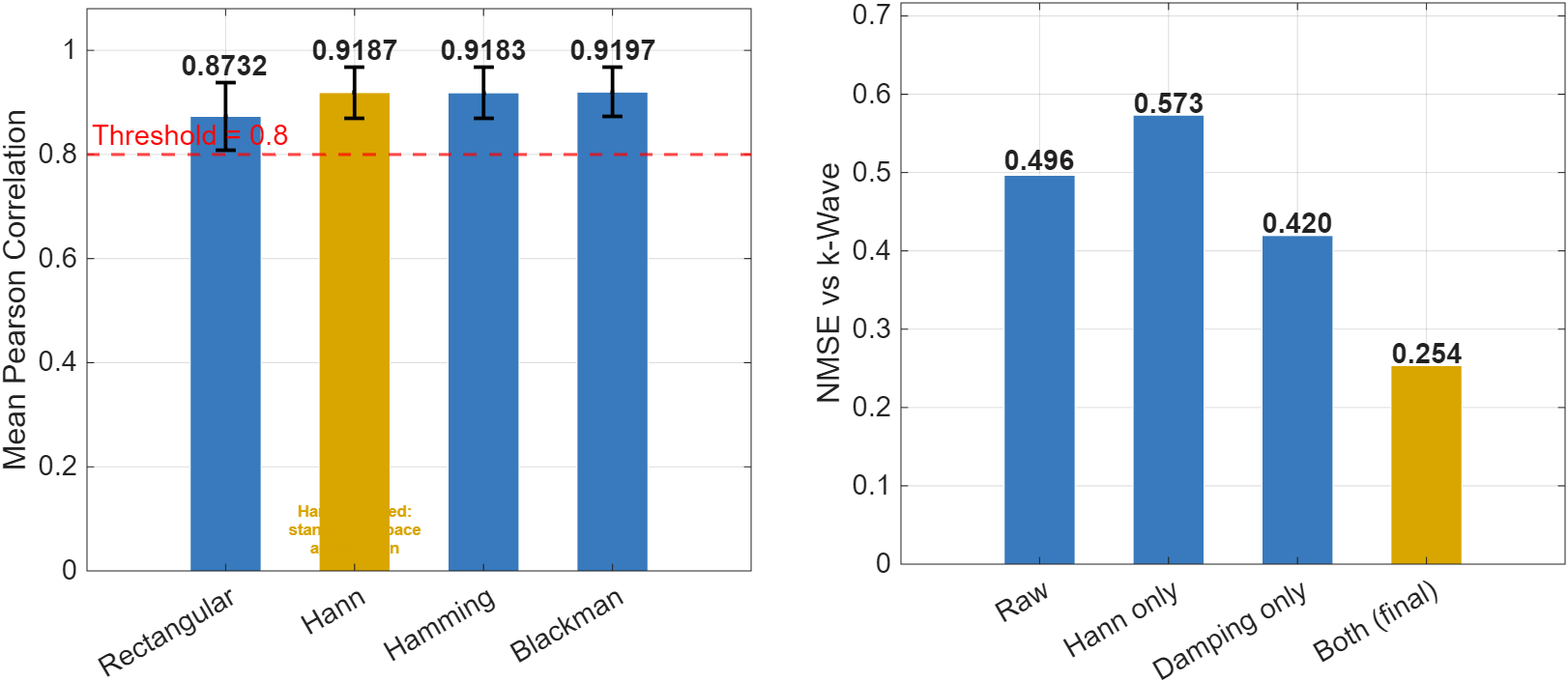}
\caption{Mean Pearson correlation achieved by each windowing techniques and the corresponding NMSE vs. K-wave.}
\label{fig5}
\end{figure}

Four apodization functions were compared: a rectangular window (i.e. no apodization), and the Hann, Hamming, and Blackman windows. The left panel of Figure \ref{fig5} shows the mean Pearson correlation achieved by each. The rectangular window produced the lowest correlation at $\bar{r}=0.8732$, confirming that unwindowed DFT matrices introduce significant Gibbs-like aliasing that degrades agreement with k-Wave. All three smooth windows improved correlation substantially and performed nearly identically; Hann: $\bar{r}=0.9187$, Hamming: $\bar{r}=0.9183$, Blackman: $\bar{r}=0.9197$, with differences of less than 0.002 between them. The Hann window was selected for all subsequent experiments as it is the conventional standard for k-space apodization in pseudospectral methods, and its performance is indistinguishable from the alternatives within the measurement noise. 

\subsubsection{Gaussian Temporal Damping (Regularisation Ablation)}\label{5.2.2}
The right panel of Figure \ref{fig5} hows an ablation study examining the individual and combined contributions of k-space apodization and Gaussian temporal damping, quantified by NMSE against k-Wave. The unregularised baseline yielded NMSE = $0.496$. Applying Hann apodization alone increased NMSE to $0.573$; a result that, while appearing counterintuitive, is consistent with the known behaviour of k-space windowing: the Hann taper reduces signal amplitude at high spatial frequencies, lowering the total signal energy and thereby inflating the energy-normalised NMSE, even though it improves the waveform shape correlation reported in Section \ref{5.2.1}. Applying Gaussian temporal damping alone ($\tau=0.9T$) reduced NMSE to $0.420$, demonstrating that late-time oscillation suppression has a more direct effect on energy-based error than apodization does. Applying both regularisations together yielded the lowest NMSE of $0.254$, a 49\% reduction from the raw baseline, confirming that the two regularisations address complementary sources of model error and act synergistically. The combined configuration was retained as the final H-matrix for all subsequent experiments.

\subsubsection{Source Position Dependence}\label{5.2.3}
The validation in Section 5.1 reported mean statistics across all sensors for a single centrally placed source. To assess whether forward model accuracy is maintained for sources at different spatial positions within the imaging domain, three source locations were evaluated: centre $(32, 32)$, off-centre $(16, 16)$, and near-edge $(8, 8)$. As shown in the left panel of Figure 6, the mean correlation remained high across all three positions, $\bar{r}=0.9907$, $0.9756$, and $0.9395$ respectively, all well above the acceptance threshold of $r_s=0.80$. The modest decrease for near-edge sources is expected, as sources close to the domain boundary generate wavefronts that interact more strongly with the PML in k-Wave, producing a slightly larger free-space mismatch. Nonetheless, the consistently high correlation across all positions confirms that the H-matrix provides a reliable forward model across the full imaging domain. 

\begin{figure}
\centering
\includegraphics[width=0.9\textwidth]{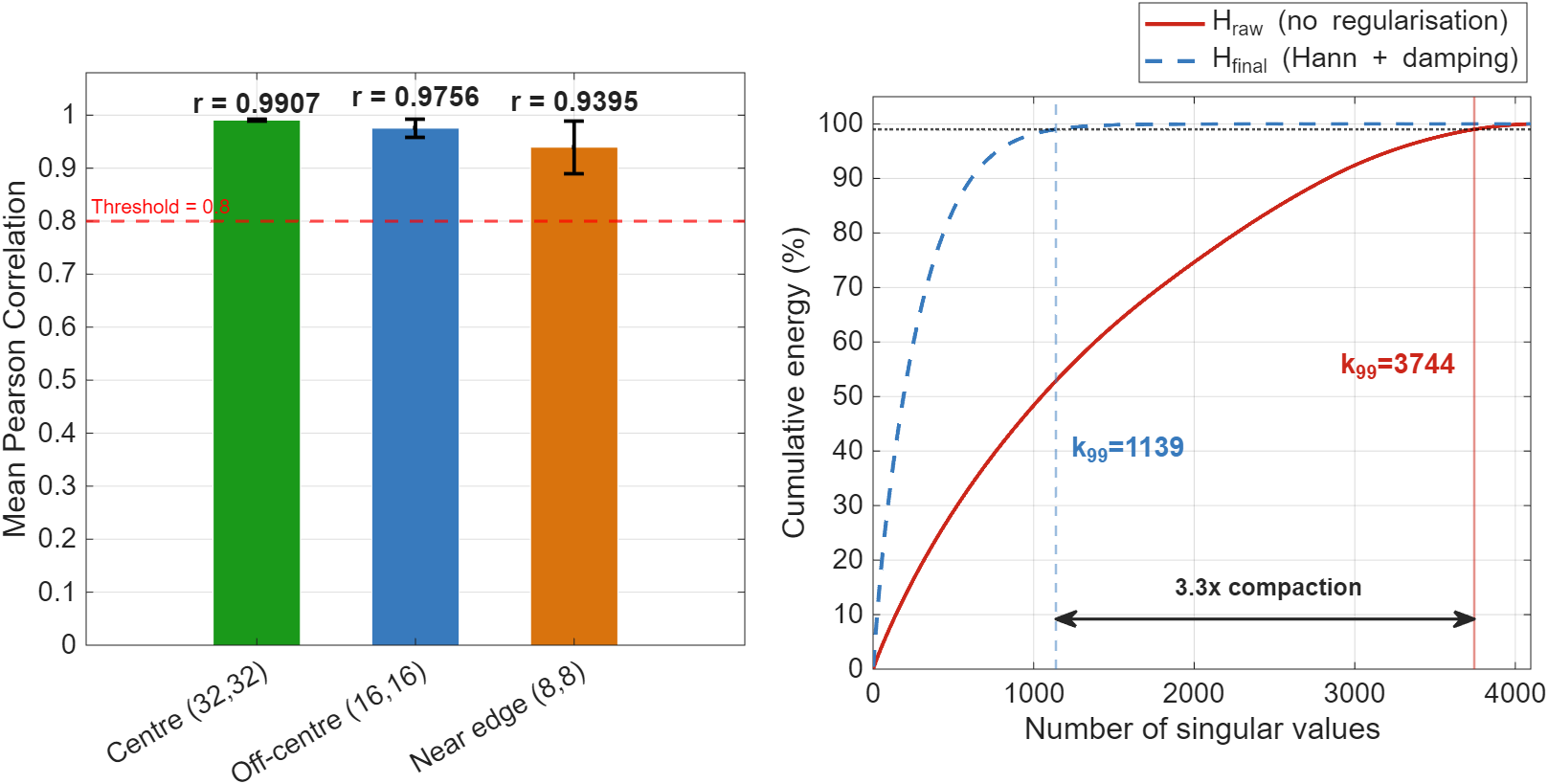}
\caption{Mean correlation for a point source in the centre, off-centre and near-edge (left); Cumulative energy versus number of singular values for $H_{\text{raw}}$ and $H_{\text{hann+dampling}}$(right).}
\label{fig6}
\end{figure}

\subsubsection{Grid Spacing Selection}

The imaging grid spacing was selected using the procedure described in
Section~\ref{3.6}, evaluating three candidate spacings,
$d_x = 0.075$, $0.100$, and $0.125$~mm, against the k-Wave reference. Agreement
was assessed using three complementary measures: the mean per-sensor Pearson
correlation, the minimum correlation across the array (a worst-case indicator),
and the fraction of sensors exceeding the acceptance threshold of $r = 0.80$.

The spacing $d_x = 0.100$~mm gave the best agreement on all three measures,
achieving a mean correlation of $r = 0.919 \pm 0.049$ with all $71$ sensors
above the $r = 0.80$ threshold. The finer $0.075$~mm grid did not improve
agreement despite its higher spatial resolution: because a smaller $d_x$
requires more time samples through the CFL condition, it lengthens the signal
window over which the free-space analytical model and the PML-bounded k-Wave
reference must agree, and the additional late-time samples are precisely where
the boundary mismatch accumulates. The coarser $0.125$~mm grid, conversely,
shortened this window but sacrificed spatial resolution without a compensating
gain in fidelity. The intermediate spacing $d_x = 0.100$~mm thus represents the
best balance between resolution and forward-model accuracy, and was adopted for
all subsequent experiments.

\subsubsection{Singular Value Compaction}\label{5.2.4}
The right panel of Figure \ref{fig6} shows the cumulative energy as a function of the number of singular values for both the raw H-matrix (H\_{raw}) and the regularised matrix (H\_{final} , Hann + damping). The number of singular values required to capture $99\%$ of the total energy, denoted $k_{99}$, decreased from 3,744 for H\_{raw}  to 1,139 for H\_{final}, representing a 3.3 compaction. This result has direct implications for compressed sensing reconstruction: a more compact H-matrix has faster effective singular value decay, which improves the conditioning of the regularised inverse problem and typically accelerates convergence of iterative solvers such as ISTA and SBTV.

\subsection{Baseline Reconstruction Methods}\label{5.3}
\begin{figure}
\centering
\includegraphics[width=1\textwidth]{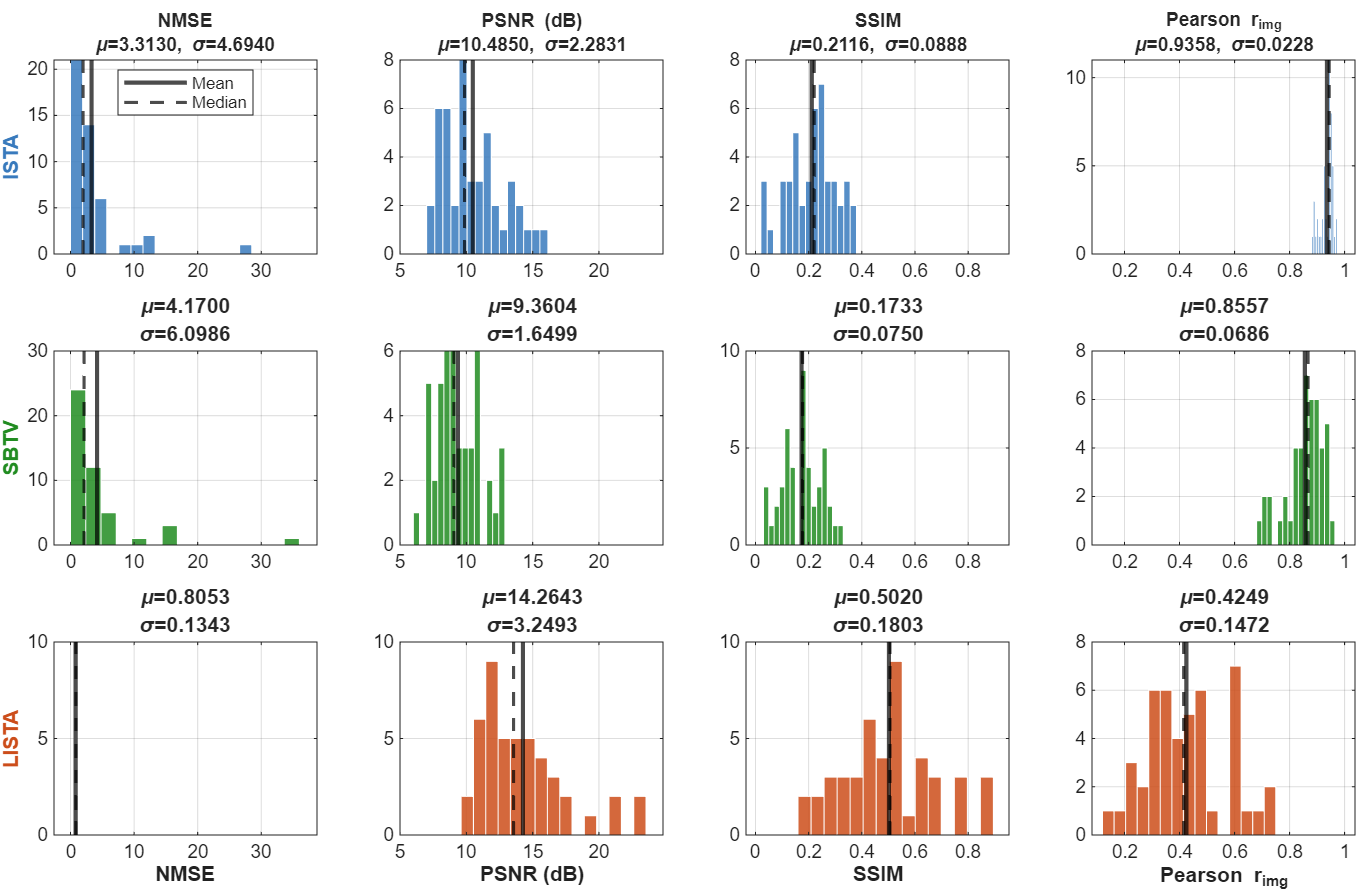}
\caption{The metric distribution histograms for all three methods (y-axis is count).}
\label{fig7}
\end{figure}

Three baseline reconstruction methods were evaluated on the held-out test set of 46 samples to establish reference performance levels against which the proposed SAN is compared. The first two methods, ISTA and SBTV, are classical iterative compressive sensing algorithms that exploit the sparsity of $\mathbf{x}_0$ and rely on repeated multiplication by $\mathbf{H}$ and ${\mathbf{H}}^{T}$  at every iteration. The third, LISTA, is a learned iterative method that unrolls a fixed number of ISTA iterations into a trainable network, replacing hand-tuned thresholds with learned parameters. All three methods use the validated H-matrix described in Section \ref{3.2} as the forward operator. Reconstruction quality is reported in terms of NMSE, PSNR, SSIM, and Pearson correlation $\bar{r}_{\text{img}}$, and the metric distributions across the test set are shown in Figure \ref{fig7}.

\subsubsection{ISTA}\label{5.3.1}
\begin{figure}
\centering
\includegraphics[width=1\textwidth]{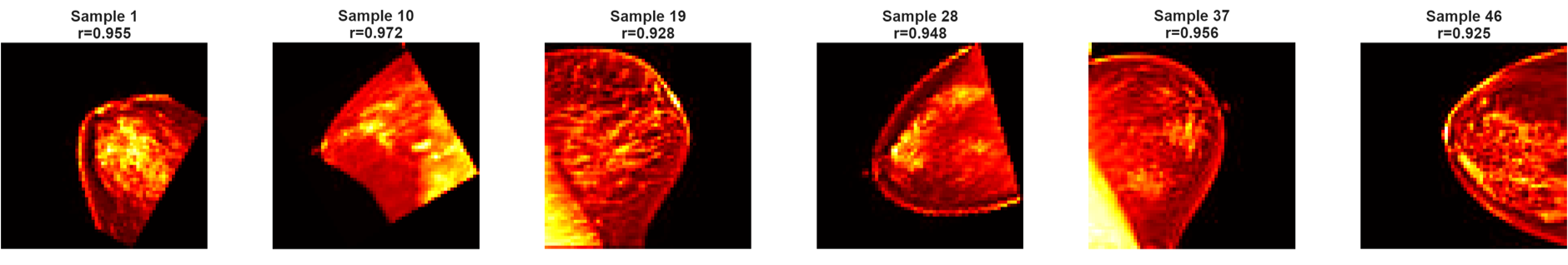}
\caption{Some examples of test images used.}
\label{fig8}
\end{figure}

\begin{figure}
\centering
\includegraphics[width=1\textwidth]{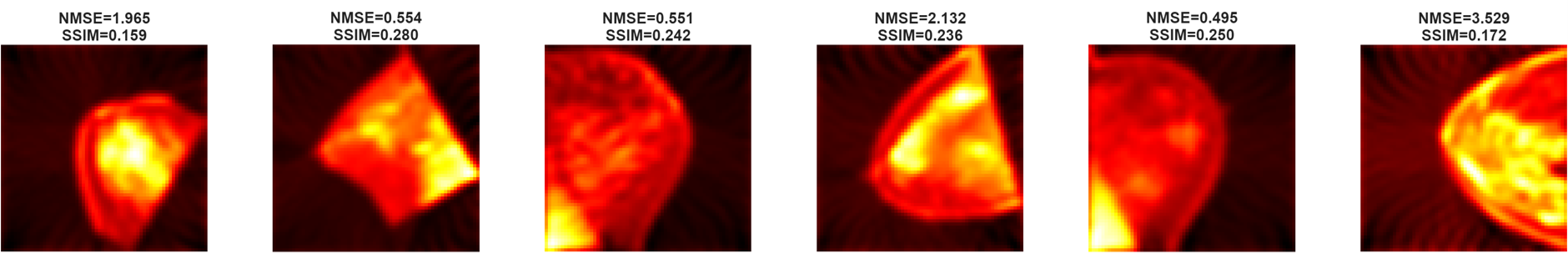}
\caption{ISTA reconstruction results.}
\label{fig9}
\end{figure}

ISTA was run with regularisation parameter $\lambda=0.009$ and a maximum of 500 iterations, at which point the cost function had converged for all test samples (final cost $2.432\times10^1$). Across the 46 test samples, ISTA achieved a mean SSIM of $0.2116\pm0.0888$, PSNR of $10.49\pm2.28$ dB, NMSE of $3.3130\pm4.6940$, and $\bar{r}_{\text{img}}$ of $0.9358\pm0.0228$. The high mean Pearson correlation indicates that the gross spatial structure of the image is recovered faithfully, but the low SSIM reflects the presence of streak artefacts and spatial smearing that degrade structural fidelity at the fine scale. The large standard deviation of the NMSE ($\sigma=4.6940$) reveals considerable variation in reconstruction quality across test samples, with a small number of samples producing substantially higher error than the majority, as visible in the heavy right tail of the NMSE histogram. Original test samples are shown in Figure \ref{fig8} and representative visual results shown in Figure \ref{fig9} confirm that ISTA recovers the coarse extent of the vessel distribution but introduces ringing artefacts along high-contrast edges and fails to resolve fine vessel detail.

\subsubsection{SBTV}\label{5.3.2}
\begin{figure}
\centering
\includegraphics[width=1\textwidth]{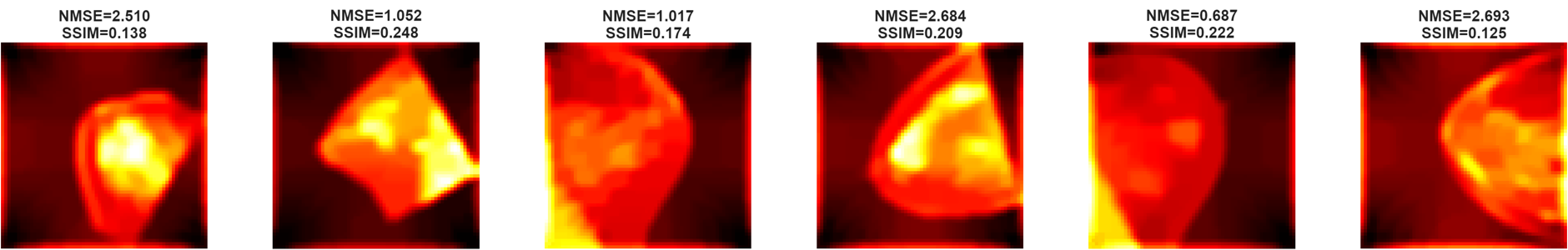}
\caption{SBTV reconstruction results}
\label{fig10}
\end{figure}

SBTV was run with the optimal parameters identified during the parameter search ($\lambda=0.1000, \mu=1.0, N_{out}=76$ outer iterations). Across the 46 test samples, SBTV achieved a mean SSIM of $0.1733\pm0.0750$, PSNR of $9.36\pm1.65$ dB, NMSE of $4.1700\pm 6.0986$, and $\bar{r}_{\text{img}}$ of $0.8557\pm0.0686$. SBTV performed below ISTA on all four metrics. The lower Pearson correlation relative to ISTA ($0.856$ vs $0.936$) suggests that the total variation regularisation, while effective at suppressing noise in standard imaging problems, introduces staircase artefacts in the vessel images used here, distorting the spatial waveform structure. The extremely large NMSE standard deviation ($\sigma=6.099$) indicates that performance is highly image-dependent, with several samples producing NMSE values exceeding 15. The test images used is shown in Figure \ref{fig8} and visual inspection of Figure \ref{fig10} confirms that SBTV reconstructions exhibit over-smoothing and loss of vessel boundary sharpness relative to ISTA. 

\subsubsection{LISTA}\label{5.3.3}
\begin{figure}
\centering
\includegraphics[width=1\textwidth]{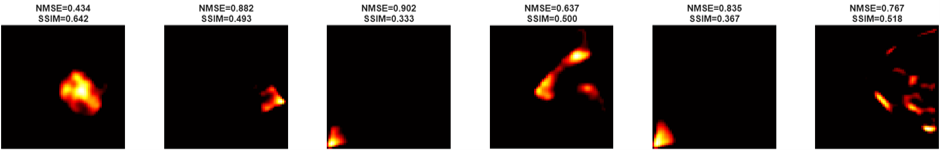}
\caption{LISTA reconstruction results}
\label{fig11}
\end{figure}

LISTA, as a learned iterative method, achieved substantially higher reconstruction quality than both classical baselines. Across the 46 test samples, LISTA achieved a mean SSIM of $0.5020\pm0.1803$, PSNR of $14.26\pm3.25$dB, NMSE of $0.8053\pm0.1343$, and $\bar{r}_{\text{img}}$ of $0.4249\pm0.1472$. The SSIM improvement over ISTA is $0.290$ in absolute terms, representing a $2.37\times$ relative gain. The lower Pearson correlation of LISTA relative to ISTA and SBTV ($0.425$ vs $0.936$ and $0.856$) is noteworthy and reflects a fundamental difference in reconstruction character: LISTA produces spatially compact, sparse reconstructions that recover the vessel locations with good structural fidelity but do not reproduce the full spatial extent of the ground truth signal in the way that the classical methods do. The test images used is shown in Figure \ref{fig8} and this trade-off is visible in Figure \ref{fig11}, where LISTA reconstructions show sharper, more localised vessel structures compared to the diffuse outputs of ISTA and SBTV, though at the cost of incomplete coverage for complex multi-vessel phantoms. The NMSE standard deviation for LISTA ($\sigma=0.1343$) is substantially lower than for ISTA and SBTV, indicating more consistent reconstruction quality across test samples. The results of all three baseline methods are summarised alongside the SAN in Table \ref{tab3}, presented in Section \ref{5.5}.

\subsection{Sensor Attention Network: Training and Validation}\label{5.4}
The SAN was trained end-to-end on the 488 augmented training samples and monitored on the 44-sample validation set. The complete network and training configuration is summarised in Table \ref{tab2}.

\begin{table}[t]
\caption{Sensor Attention Network configuration}
\label{tab2}
\begin{tabular*}{\tblwidth}{@{\extracolsep{\fill}}llp{6cm}@{}}
\toprule
\textbf{Parameter} & \textbf{Symbol} & \textbf{Value} \\
\midrule
Token embedding dimension      & $d_{\mathrm{model}}$  & 128 \\
Number of attention heads      & $A$  & 4 \\
Encoder blocks                 & $L$  & 2 \\
Feed-forward hidden dimension  & $d_{\mathrm{ff}}$     & 256 \\
Vessel-weighting threshold     & --                    & 0.10 \\
Vessel weight                  & $w_{\mathrm{vessel}}$ & 8 \\
Background weight              & $w_{\mathrm{bg}}$     & 1 \\
Optimiser                      & -- & Adam ($\beta_1=0.9,\beta_2=0.999$) \\
Initial learning rate          & -- & $3\times10^{-4}$ \\
LR decay schedule              & -- & $\times 0.5$ every 80 epochs \\
Batch size                     & -- & 8 \\
Total epochs                   & -- & 300 \\
Trainable parameters           & -- & $\approx 18\times10^{6}$ \\
\bottomrule
\end{tabular*}
\end{table}

All weights were initialised using He initialisation \cite{he2015delving}. The vessel-weighted MSE loss of equation (\ref{eq:16}) was minimised with Adam, with the learning rate halved every 80 epochs. The checkpoint with the lowest validation loss across the 300-epoch schedule was retained for evaluation.
\subsubsection{Test-set performance}\label{5.4.1}
The trained network was evaluated on the 46-sample held-out test set, which was not used at any stage of training or model-selection. Across the test set, the SAN achieved a mean SSIM of $0.522 \pm 0.225$, a mean PSNR of $22.09 \pm 7.42$ dB, a mean NMSE of $0.233 \pm 0.172$, a mean {$\bar{r}_{\text{img}}$ of $0.836 \pm 0.103$, and a mean coefficient of determination $R^2$  of $0.654 \pm 0.237$. Of the 46 test samples, 21 ($46\%$) achieved SSIM > 0.5 and 10 ($22\%$) achieved SSIM > 0.7, with the best sample reaching SSIM = $0.965$. The standard deviation of 0.225 in SSIM reflects substantial sample-to-sample variation: the network performs strongly on phantoms whose vessel layout is well-represented in the training distribution and less well on outlier topologies, as visible in the lower tail of the distribution in Figure \ref{fig12}.

\begin{figure}
\centering
\includegraphics[width=0.6\textwidth]{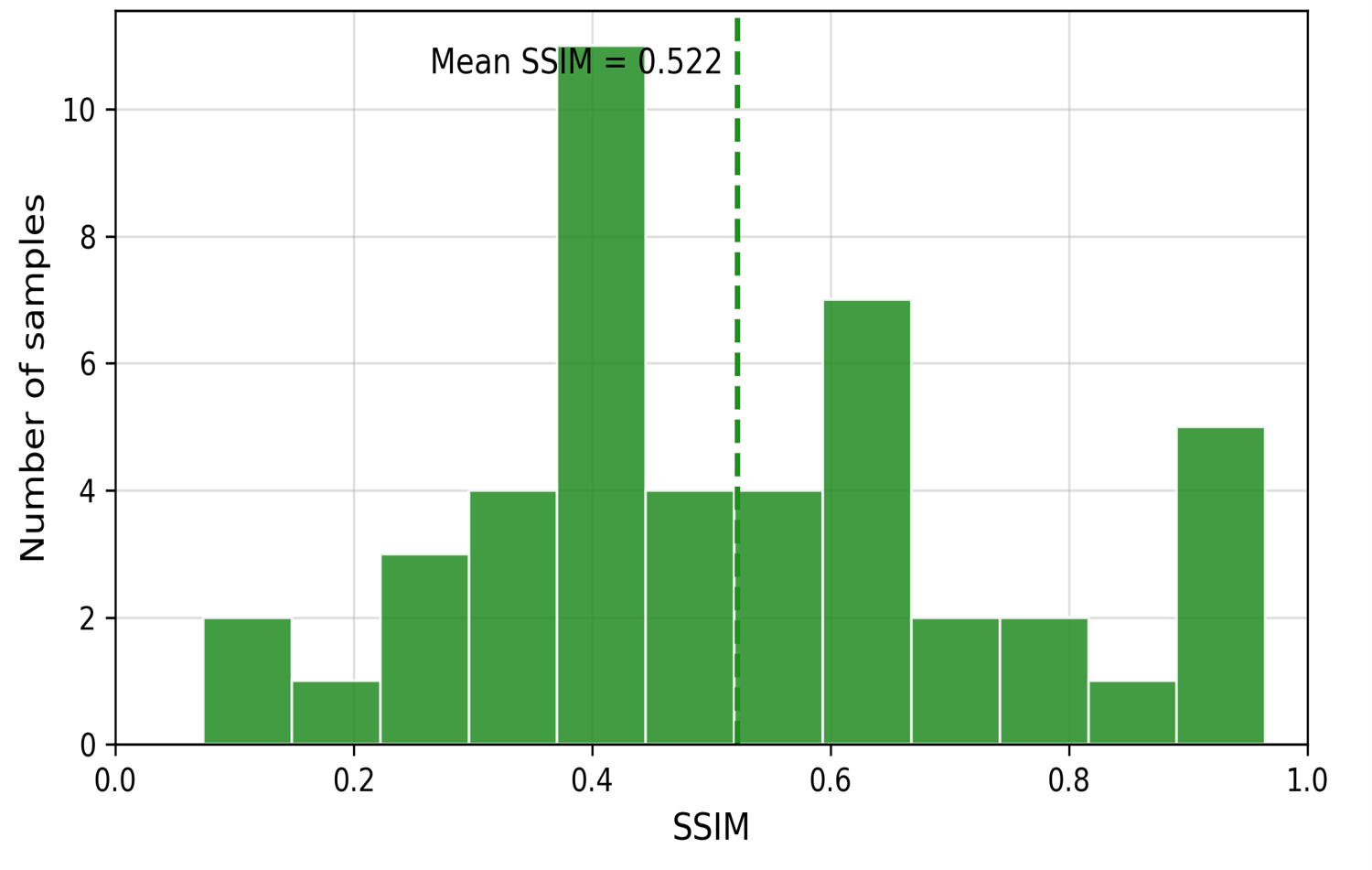}
\caption{Per-sample SSIM distribution of the SAN reconstructions on the 46-sample test set. Dashed vertical line marks the mean.}
\label{fig12}
\end{figure}

\subsubsection{Qualitative results}\label{5.4.2}
Figure \ref{fig13} shows reconstructions for six randomly selected test samples (top row: ground truth; bottom row: SAN prediction). The network recovers the position, orientation, and branching topology of the vessel structures faithfully across all six cases, with sharp boundaries and well-localised reconstructions. Minor amplitude attenuation is observed at the thinnest vessel segments - a consequence of the limited spatial-frequency content of the training distribution and the smoothing effect of the sigmoid output activation - but the geometric fidelity remains high. 
Beyond reconstruction quality, the proposed network offers a substantial inference-cost advantage. Each reconstruction is produced in a single forward pass through the trained network, with no H-matrix multiplication required, and completes in a few milliseconds on a single mid-range GPU. By contrast, the iterative baselines reported in Section V-C require either 200 (ISTA), 4 outer (SBTV), or 16 (LISTA) matrix-vector products of cost $O(MN) \approx 4.65\times10^7 $floating-point operations each, leading to inference times that are at least one order of magnitude longer.

\begin{figure}
\centering
\includegraphics[width=1\textwidth]{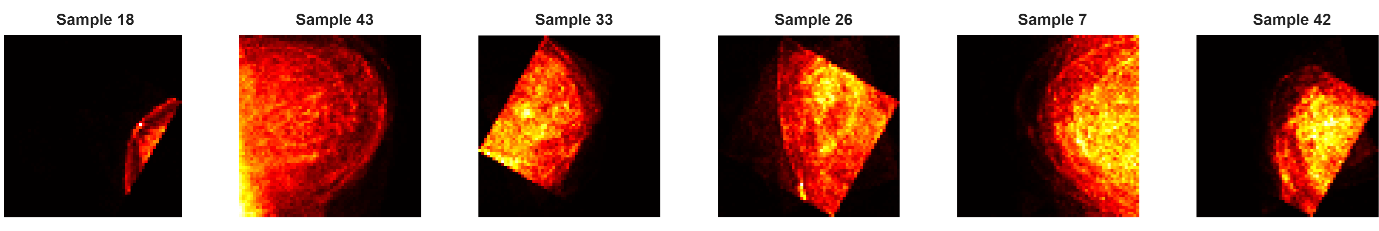}
\caption{Six random test-set reconstructions}
\label{fig13}
\end{figure}

\subsection {SAN versus baselines}\label{5.5}
This section presents a direct comparison of the four reconstruction methods evaluated in this work - ISTA, SBTV, LISTA, and the proposed SAN - on the same 46-sample held-out test set, using the validated H-matrix described in Section 3.2 as the forward operator. All metrics are computed on images normalised to the range [0, 1]. Aggregate results are summarised in Table \ref{tab3}.

\begin{table*}[t]
\caption{Reconstruction performance of all four methods on the 46-sample test set. Mean $\pm$ standard deviation. The best value in each column is shown in bold; ties or non-significant differences are not bolded.}
\label{tab3}

\centering
\setlength{\tabcolsep}{10pt}
\renewcommand{\arraystretch}{1.2}

\begin{tabular}{@{}lcccc@{}}
\toprule
\textbf{Method} & \textbf{SSIM} & \textbf{PSNR (dB)} & \textbf{NMSE} & \textbf{$\bar{r}_{\text{img}}$} \\
\midrule
ISTA &
$0.212 \pm 0.089$ &
$10.49 \pm 2.28$ &
$3.313 \pm 4.694$ &
$\mathbf{0.936 \pm 0.023}$ \\

SBTV &
$0.173 \pm 0.075$ &
$9.36 \pm 1.65$ &
$4.170 \pm 6.099$ &
$0.856 \pm 0.069$ \\

LISTA &
$0.502 \pm 0.180$ &
$14.26 \pm 3.25$ &
$0.805 \pm 0.134$ &
$0.425 \pm 0.147$ \\

\textbf{SAN (proposed)} &
$\mathbf{0.522 \pm 0.225}$ &
$\mathbf{22.09 \pm 7.42}$ &
$\mathbf{0.233 \pm 0.172}$ &
$0.836 \pm 0.103$ \\
\bottomrule
\end{tabular}
\end{table*}

The SAN achieves the highest mean SSIM and PSNR, and the lowest mean NMSE, of the four methods. To establish whether these differences are statistically significant given the modest test-set size ($n = 46$), each pairwise comparison was assessed using both a paired t-test on the per-sample metric values and a Wilcoxon signed-rank test, the latter providing a non-parametric check that does not assume normally distributed differences. Results are summarised in Table \ref{tab4}

\begin{table*}[t]
\caption{Paired statistical comparison of the SAN against each baseline on the test set ($n=46$). The mean difference is computed as (SAN - baseline) for SSIM, PSNR and Pearson $\bar{r}_{\text{img}}$ (where higher is better) and as (baseline - SAN) for NMSE (where lower is better), so that a positive value always favours the SAN. Significance codes: *** $p<0.001$, ** $p<0.01$, * $p<0.05$, not significant (n.s.) $p\geq0.05$.}
\label{tab4}

\centering
\footnotesize
\setlength{\tabcolsep}{5pt}
\renewcommand{\arraystretch}{1.15}

\begin{tabular}{llllll}
\toprule
\textbf{Comparison} &
\textbf{Metric} &
\textbf{Mean diff} &
\textbf{95\% CI} &
\textbf{p (t-test)} &
\textbf{p (Wilcoxon)} \\
\midrule

SAN vs ISTA & SSIM      & $+0.310$ & $$[+0.229, +0.392]$$ & $1.1\times10^{-9}$ ***  & $1.2\times10^{-7}$ *** \\
SAN vs ISTA & PSNR (dB) & $+11.60$ & $[+9.21, +14.00]$  & $1.1\times10^{-12}$ *** & $3.5\times10^{-9}$ *** \\
SAN vs ISTA & NMSE      & $+3.08$  & $[+1.69, +4.47]$   & $5.2\times10^{-5}$ ***  & $3.5\times10^{-9}$ *** \\
SAN vs ISTA & Pearson $\bar{r}_{\text{img}}$ & $-0.100$ & $[-0.129, -0.071]$ & $1.5\times10^{-8}$ *** & $4.1\times10^{-7}$ *** \\

SAN vs SBTV & SSIM      & $+0.349$ & $[+0.270, +0.427]$ & $1.6\times10^{-11}$ *** & $1.4\times10^{-8}$ *** \\
SAN vs SBTV & PSNR (dB) & $+12.73$ & $[+10.41, +15.05]$ & $2.2\times10^{-14}$ *** & $3.5\times10^{-9}$ *** \\
SAN vs SBTV & NMSE      & $+3.94$  & $[+2.14, +5.74]$   & $6.6\times10^{-5}$ ***  & $3.5\times10^{-9}$ *** \\
SAN vs SBTV & Pearson $\bar{r}_{\text{img}}$ & $-0.020$ & $[-0.051, +0.011]$ & $0.20$ (n.s.) & $0.057$ (n.s.) \\

SAN vs LISTA & SSIM      & $+0.020$ & $[-0.046, +0.086]$ & $0.55$ (n.s.) & $0.90$ (n.s.)\\
SAN vs LISTA & PSNR (dB) & $+7.83$  & $[+5.64, +10.02]$  & $5.2\times10^{-9}$ ***  & $4.0\times10^{-9}$ *** \\
SAN vs LISTA & NMSE      & $+0.572$ & $[+0.506, +0.638]$ & $1.1\times10^{-21}$ *** & $4.0\times10^{-9}$ *** \\
SAN vs LISTA & Pearson $\bar{r}_{\text{img}}$ & $+0.411$ & $[+0.360, +0.462]$ & $2.3\times10^{-20}$ *** & $3.5\times10^{-9}$ *** \\

\bottomrule
\end{tabular}
\end{table*}

Three observations follow from this analysis. First, the SAN dominates the classical iterative baselines (ISTA and SBTV) on every fidelity metric. The improvements in SSIM ($ +0.31$ over ISTA, $+0.35$ over SBTV), PSNR ($+11.6$ dB and $+12.7$ dB), and NMSE ($3.1$ and $3.9$ lower respectively) are all highly statistically significant ($p < 10^{-4}$) under both the parametric and non-parametric tests. The single exception is Pearson correlation, where ISTA retains a small but significant lead ($\Delta \bar{r}_{\text{img}} = -0.10, \; p < 10^{-8}$) and SBTV is statistically indistinguishable from the SAN ($p = 0.20$; not significant (n.s.)). The high Pearson correlation of ISTA reflects its well-known behaviour on sparse signals: the algorithm recovers the gross spatial layout of the vessels with strong linear similarity to the ground truth while producing diffuse, low-amplitude reconstructions whose pixel-level structural fidelity (SSIM = 0.21) is poor. The SAN, by contrast, achieves only a modestly lower Pearson $\bar{r}_{\text{img}}$ yet a substantially higher SSIM and PSNR, indicating reconstructions that are structurally faithful as well as numerically accurate.
Second, the comparison against LISTA, the strongest baseline establishes the proposed network's principal contribution. On SSIM the two methods are statistically indistinguishable ($\Delta = +0.020, \; p = 0.55 {\text{(n.s.}})$), reflecting that both produce structurally compact reconstructions of comparable quality at the single-scale SSIM level. On every other metric, however, the SAN is decisively superior: PSNR is $7.83$ dB higher ($p <10^{-8}$), NMSE is $3.5 \times$ lower ($p < 10^{-21}$), and Pearson $\bar{r}_{\text{img}}$ is $0.41$ higher ($p < 10^{-20}$). The lower Pearson $\bar{r}_{\text{img}}$  of LISTA is consistent with a known limitation of unrolled networks trained with limited data: the learned proximal operator applies too strong a sparsifying threshold at each iteration, leaving incomplete coverage on multi-vessel phantoms that the SAN reconstructs more completely.
Third, the qualitative comparison can be made by viewing the SAN reconstructions in Figure \ref{fig13} above alongside the ISTA, SBTV, and LISTA reconstruction examples already presented in Section \ref{5.3}. ISTA and SBTV recover the coarse vessel distribution but exhibit smearing and ringing along high-contrast edges, with SBTV additionally producing piecewise-constant 'staircase' artefacts characteristic of total-variation regularisation. LISTA produces sharper, more spatially compact reconstructions than the classical methods but tends to miss vessel segments far from the dominant signal centroid, leaving incomplete coverage on multi-vessel phantoms, visible quantitatively as the substantially lower Pearson correlation despite high SSIM. The SAN reconstructions in Figure \ref{fig13} match the ground-truth vessel layout closely on most samples, with sharp boundaries and faithful branching topology. Failure modes are restricted to phantoms whose vessel patterns lie outside the augmented training distribution, corresponding to the lower tail of the SSIM histogram in Figure \ref{fig12}.
In summary, the proposed SAN produces reconstructions of significantly higher quality than the classical iterative methods (ISTA, SBTV) on every fidelity metric, and matches or exceeds the strongest learned baseline (LISTA) on every metric except SSIM, where the two are statistically indistinguishable. This is achieved while bypassing the H-matrix entirely at inference time, providing an order-of-magnitude reduction in reconstruction time relative to all three iterative baselines.

\section{DISCUSSION}\label{6} 
The principal finding of this work is that a Transformer-based sensor-as-token architecture can produce PAT reconstructions of significantly higher fidelity than the classical iterative baselines (ISTA and SBTV) and at least comparable to the strongest unrolled-network baseline (LISTA), while bypassing the H-matrix at inference. The paired statistical comparison in Section 5.5  places this claim on rigorous footing: SAN attains improvements of $+11.6$ dB PSNR over ISTA, $+12.7$ dB over SBTV, and $+7.83$ dB over LISTA, with each comparison significant at $p<10^{-8}$ under both parametric (paired t-test) and non-parametric (Wilcoxon signed-rank) tests. The NMSE reduction relative to LISTA, the strongest baseline, is a factor of $3.5 \times (p<10^{-21})$, establishing the proposed network as a clear quantitative improvement over the prior state of the art at the test-set scale studied here \cite{john2023advancing10}.

\subsection{Interpretation of the Reconstruction Trade-Offs}\label{6.1}
Three observations from Section \ref{5.5} warrant further discussion. First, the Pearson correlation coefficient retains a small advantage for ISTA ($0.936$ vs SAN’s $0.836$, $\Delta \bar{r}_{img}=-0.10, \;p<10^{-8}$). This is consistent with the well-known behaviour of $l_{1}$-regularised iterative solvers on sparse signals \cite{john2022fast}: such methods recover the gross spatial layout of the vessels with high linear similarity to the ground truth, but the resulting reconstructions are diffuse and low-amplitude, which depresses the single-scale SSIM ($0.21$ for ISTA). The SAN, by contrast, achieves only a modestly lower Pearson $\bar{r}_{\text{img}}$ but a substantially higher SSIM ($0.522$), reflecting reconstructions that are simultaneously structurally faithful at the pixel level and numerically accurate in the energy sense.
Second, on the single-scale SSIM the SAN and LISTA are statistically indistinguishable ($\Delta=+0.020,\; p=0.55$ (n.s.). This is itself an informative result: SSIM is sensitive to fine boundary structure, and LISTA, being a learned variant of ISTA, inherits its ability to produce sparse, well-localised reconstructions. The decisive separation between the two methods appears in the energy-based metrics (PSNR, NMSE) and in Pearson correlation, where SAN’s coverage of the full vessel extent and amplitude is substantially better than LISTA’s. This pattern is consistent with the known limitation of unrolled networks trained on limited data: the learned proximal operator can become overly aggressive at sparsifying the iterate, leaving incomplete coverage on multi-vessel phantoms.
Third, the multi-scale structural similarity index (MS-SSIM) was deliberately excluded from the principal comparison table. On the same test set, ISTA attained a higher MS-SSIM ($0.866$) than the SAN ($0.794$) despite its much lower single-scale SSIM. This counterintuitive result is a consequence of MS-SSIM’s weighting of coarse-scale structural similarity, which is well preserved by diffuse low-amplitude reconstructions such as those produced by ISTA. Since fine-scale boundary structure is the metric of greatest clinical interest for vessel imaging, the single-scale SSIM is the more discriminating measure for this task, and the MS-SSIM is reported only as a SAN-specific descriptor in Section 5.4.

\subsection{Computational Efficiency at Inference}\label{6.2}
The computational advantage of the SAN at inference is independent of the reconstruction-quality argument and is therefore worth separating. Each iterative baseline performs at least one matrix–vector product with $\bf{H}$ per iteration, with cost $O(MN) \approx 4.65\times10^7$ floating-point operations for the chosen grid size of $64\times64$ pixels and $N_sN_t=11,360$ measurements. ISTA uses 200 such iterations per reconstruction; SBTV uses 60 outer plus 5 inner conjugate-gradient iterations; LISTA uses 16 unrolled layers. The SAN, by contrast, performs a single forward pass dominated by dense matrix multiplications totalling on the order of $10^7$ operations, completing in 1 to 5 ms on a single mid-range GPU. This represents at least an order-of-magnitude reduction in inference latency relative to all three iterative baselines, and brings PAT reconstruction within the latency budget of real-time clinical workflows.

\subsection{Limitations}\label{6.3}
Three limitations of the present study deserve explicit acknowledgement. First, the training corpus comprises 191 phantoms drawn from a single public source (Radiopaedia), expanded to 488 augmented samples by small spatial translations. While the held-out test set was strictly disjoint and the validation set was used only for checkpoint selection, the absolute generalisation capacity of the network to phantoms outside this distribution remains unproven. Failure cases observed in the lower tail of the SSIM histogram (Figure \ref{fig12}) correspond predominantly to phantoms with vessel topologies poorly represented in the training distribution.
Second, the analytical H-matrix is constructed under the assumption of a homogeneous, lossless acoustic medium with constant speed of sound. Real biological tissue exhibits acoustic heterogeneity, frequency-dependent attenuation, and in some cases multiple scattering, none of which are modelled in the present forward operator. Although the validation against k-wave confirms that the analytical model faithfully reproduces idealised forward physics, the gap between the analytical training distribution and experimental measurements must be quantified before clinical translation.
Third, the present study reports results for a single grid spacing ($d_x=0.100$ mm), sensor configuration ($N_s=71$ on a square boundary), and image size ($64\times64$). The scaling of SAN performance with image resolution, sensor count, and grid spacing has not been characterised here; this is the subject of ongoing work.
\subsection{Comparison to Prior Self-Work}\label{6.4}
The proposed architecture is positioned as a complement, not a replacement, of the unrolled-network methods previously reported by the present authors [10], [14]. Unrolled networks remain valuable for problems in which the forward operator H changes between samples (e.g., different sensor geometries, patient-specific scan protocols) and physical interpretability of the per-layer iterate is required for clinical acceptance. The SAN is most appropriate when the sensor geometry is fixed across the patient population and inference latency is the dominant constraint. Both approaches benefit from the validated analytical H-matrix introduced in Section 3, which can be used for training data generation regardless of the downstream reconstruction architecture.

\section{CONCLUSION AND FUTURE WORK}\label{7}
This paper introduced the Sensor Attention Network (SAN), a Transformer-based architecture for photoacoustic tomography image reconstruction that maps raw sensor measurements directly to the reconstructed image without invoking the forward operator at inference. The contributions are threefold: (i) an analytical k-space H-matrix was constructed and validated against the k-Wave pseudo-spectral solver under matched geometry, achieving a mean per-sensor Pearson correlation of $\bar r=0.919 \pm0.049$, with two targeted regularisations (Hann apodization and Gaussian temporal damping) reducing the energy-normalised mismatch by 49\%; (ii) the SAN architecture was introduced and trained end-to-end on 488 augmented PAT phantoms; and (iii) a rigorous statistical comparison against ISTA, SBTV, and LISTA on a 46-sample held-out test set established that the SAN achieves the highest mean SSIM (0.522), the highest mean PSNR (22.09 dB), and the lowest mean NMSE (0.233), with paired t-tests and Wilcoxon signed-rank tests confirming significance at $p<10^{-8}$ for the principal claims. By eliminating the per-iteration $\mathbf{Hx}$ multiplication at inference, the SAN reduces reconstruction latency by at least an order of magnitude relative to all three iterative baselines.
Several directions for future work are identified. First, the training corpus should be expanded with more diverse vessel topologies and, where possible, with experimentally acquired sensor signals from phantom and ex-vivo tissue measurements, in order to characterise the network’s generalisation capacity beyond the simulated regime \cite{john2023advancing10}. Second, the data augmentation strategy can be made more physically faithful by applying the corresponding sensor-index permutation when rotating or reflecting the phantom, a procedure that is exact for a sensor array with discrete rotational symmetry and that avoids the analytical/numerical distribution mismatch encountered in preliminary augmented experiments. Third, the SAN architecture should be extended to three-dimensional PAT by replacing the two-dimensional spatial decoder with a 3-D volumetric decoder, and by accommodating cylindrical or hemispherical sensor geometries common in clinical scanners \cite{huynh2026feasibility30}. Moreover, a learned MLP encoding of the physical sensor coordinates can be considered as a future work. Fourth, the integration of physics-informed loss terms, penalising deviation of the reconstructed image from the analytical forward model under perturbed sensor inputs, offers a principled means of combining the data efficiency of unrolled networks with the inference speed of the direct architecture proposed here. Fifth, the present design should be evaluated on heterogeneous acoustic media and frequency-dependent attenuation, possibly with a learned correction layer to bridge the gap between the homogeneous analytical training distribution and clinical measurements. Finally, sixth, the network’s transferability to related tomographic modalities, such as thermoacoustic tomography and ultrasound computed tomography, merits investigation, as the underlying sensor-as-token paradigm is not specific to PAT and may generalise to any inverse problem in which a fixed sensor array produces a sparse-view measurement vector.

\section*{Declaration of competing interest}
The authors declare that they have no known competing financial
interests or personal relationships that could have appeared to influence
the work reported in this paper.






\end{document}